\documentclass[journal]{IEEEtran}
\usepackage{amssymb}
\usepackage{amsmath}
\usepackage{mathrsfs}
\usepackage{amsthm}
\usepackage{setspace}
\usepackage{booktabs}
\usepackage{amsfonts}
\usepackage{subfigure}

\usepackage{multirow}
\usepackage{graphicx}

\usepackage{float}
\usepackage{subeqnarray}

\usepackage{cases}

\ifCLASSINFOpdf
\else
\fi
\makeatletter
\thm@headfont{\sc}
\makeatother
\newtheorem{theorem}{Theorem}
\newtheorem{lemma}[theorem]{Lemma}

\begin{document}
%
\title{Efficient Background Modeling Based on Sparse Representation and Outlier Iterative Removal}
%
%
%

\author{Linhao~Li,
        Ping~Wang,
        Qinghua~Hu,~\IEEEmembership{Senior~Member,~IEEE},
        ~Sijia~Cai

\thanks{This work was partially supported by the National Natural Science Foundation of China (No. 51275348, No. 61379014, No. 6122210, and No. 61432011), and New Century Excellent Talents in University (Grant No. NCET-12-0399).}
\thanks{L. Li is with the School of Computer Science and Technology, Tianjin University,  Tianjin 300072, P.R. China. (e-mail: lilinhao@tju.edn.cn).}
\thanks{P. Wang is with the School of Science and the Center for Applied Mathematics of Tianjin University, Tianjin University,  Tianjin 300072, P.R. China. (e-mail: wang\underline{ }ping@tju.edu.cn).}
\thanks{Q. Hu is with the School of Computer Science and Technology and the Tianjin Key Laboratory of Cognitive Computing and Application, Tianjin University,  Tianjin 300072, P.R. China. (e-mail: huqinghua@tju.edu.cn).}
\thanks{S. Cai is with the School of Science, Tianjin University,  Tianjin 300072, P.R. China. (e-mail: caisj@tju.edu.cn).}}

%
%

\markboth{Journal of \LaTeX\ Class Files,~Vol.~11, No.~4, December~2012}%
{Shell \MakeLowercase{\textit{et al.}}: Bare Demo of IEEEtran.cls for Journals}
%



\maketitle

\begin{abstract}
Background modeling is a critical component for various vision-based applications. Most traditional methods tend to be inefficient when solving large-scale problems. In this paper, we introduce sparse representation into the task of large-scale stable-background modeling, and reduce the video size by exploring its "discriminative" frames. A cyclic iteration process is then proposed to extract the background from the discriminative frame set. The two parts combine to form our Sparse Outlier Iterative Removal (SOIR) algorithm. The algorithm operates in tensor space to obey the natural data structure of videos. Experimental results show that a few discriminative frames determine the performance of the background extraction. Further, SOIR can achieve high accuracy and high speed simultaneously when dealing with real video sequences. Thus, SOIR has an advantage in solving large-scale tasks.
\end{abstract}

\begin{IEEEkeywords}
Background Modeling; Sparse Representation; Tensor Analysis; Principal Component Pursuit; Alternating Direction Multipliers Method; Markov Random Field.
\end{IEEEkeywords}

%
\IEEEpeerreviewmaketitle

\section{Introduction}
\label{sec1}
The background modeling of a video sequence is a key part in many vision-based applications such as real-time tracking \cite{HCQ2011,KLM2012}, information retrieval, and video surveillance \cite{T2011,AOM2006}. In a video sequence, some scenes will remain nearly constant, even though they may be polluted by noise \cite{SHL2012}. The invariable aspect is the background. A model for extracting the background is an important tool that can help us  handle a video sequence, especially one taken in a public area \cite{DH2013}. Background modeling is also an essential step in many foreground detection tasks \cite{WJW2008,MO2012,VN2008}. Once the background is extracted, we can detect or even track the foreground information by comparing a new frame with the learned background model \cite{AOM2006}.

There are two challenges to a background modeling algorithm. First, although we consider the background to be stationary, it is often interfered with by certain factors such as fluttering flags, waving leaves, or rippling water \cite{YCCY2007}. In addition, other issues such as signal noise, sudden lighting variations, and shadows \cite{STS2011, AT2008}, may prevent us from distinguishing the background from a video sequence. Second, the data on practical problems is increasing with the development of new technologies and improvements in the equipment used. However, there is also an increasing demand for efficient background modeling techniques, and fast tracking of massive video sequences is required for certain practical tasks, like crime detection and recognition. As a result, it has become an urgent task to develop an efficient and robust algorithm for practical background modeling.

A large number of background modeling methods have been reported in the literature over the past decades. Most researchers have regarded a series of pixel values as features, and set up pixel-wise models. Initially, each pixel-value series is modeled using a Gaussian distribution, e.g., the Single Gaussian (SG) model developed in 1997 \cite{CATA1997} and the Multiple of Gaussian (MOG) model developed in 1999 \cite{CW1999}. Some improved Gaussian-based algorithms \cite{D2005, JHGJ2011, HJT2011} also achieved a high level of performance in the few years following the release of these above models. In addition, clustering methods have also been used to model a background, e.g., codebook \cite{JYCC2011, AM2011} and time-series clustering \cite{AN2012}. Furthermore, a non-parametric method was proposed in 2000 \cite{ADL2000} and improved in 2012 \cite{EMA2012}, and has shown a competitive performance. The Visual Background Extractor (ViBe) was recently  proposed in 2012 and later improved, and performs better than most popular pixel-wise techniques \cite{MO2012,OM2011}. These methods solve the background problem by building a model for each pixel and initializing the models during the training process. High accuracy is obtained if sufficient training data are provided, but more training data means additional training time.

Another type of modeling techniques is to set up the model at the region level. Some works have focused on the local region, and different local features have been proposed \cite{STS2011, LWIQ2004, LWXG2010,SGV2010}. In addition, global-region based algorithms have also been proposed. In 2000, Oliver et al. \cite{NBA2000} first modeled a background using a Principal Component Analysis (PCA), i.e., they modeled the background by projecting high-dimensional data into a lower dimensional subspace. Robust PCA developed in 2010 \cite{ZXJEY2010}, and Principal Component Pursuit (PCP) developed in 2011 \cite{EXYJ2011}, have shown their superiority over the original PCA. Based on these models, heuristic background methods have also been introduced \cite{GA2011,CTE2012,XCW2013}. These PCA-based models omit the training process and use data to extract the background directly. However, Singular Value Decomposition (SVD) is an inevitable time-consuming step in a PCA-based model, and thus these models are limited in large-scale tasks because their speed and memory requirements are all sensitive to the scale of the data.

Sparse representation and dictionary learning is also an important region-based method. It is widely employed in the tasks of computer vision such as face recognition \cite{YDJ2011,JAASY2009,EGR2012}, classification \cite{JAASY2009,WDLXD2013} and denoising \cite{JYW2009}. Some researchers have introduced sparse representation into background modeling \cite{RAM2011,CXW2011}. They modeled the background by the dictionary and regarded the foreground as noise. Besides, they made some assumptions of independence among different pixels. However, these assumptions fail in many practical tasks where the foreground region is usually not sparse and some pixels are highly correlated.

In this paper, first, we use sparse representation to reduce the video size by exploring the "discriminative" frames of the video, instead of modeling the background directly. No assumption is needed in this process. We then extract the background from these discriminative frames using a PCP-based cyclic iteration. These two steps combine to form our algorithm, i.e., Sparse Outlier Iterative Removal (SOIR) algorithm. The framework of the algorithm is shown in Fig. 1.

SOIR meets the demand of global-region based background models on solving large-scale problems. For our algorithm, we rebuild the PCP model based on a rank-1 hypothesis. Moreover, our algorithm operates in a tensor space in order to obey the natural data structure of the videos. We detect foreground objects using the Markov Random Field (MRF) once the background is extracted. Experimental results show that our algorithm can achieve high accuracy and high speed simultaneously when dealing with real-life video sequences.

\begin{figure}[H]
\begin{center}
\includegraphics[width=0.43\textwidth]{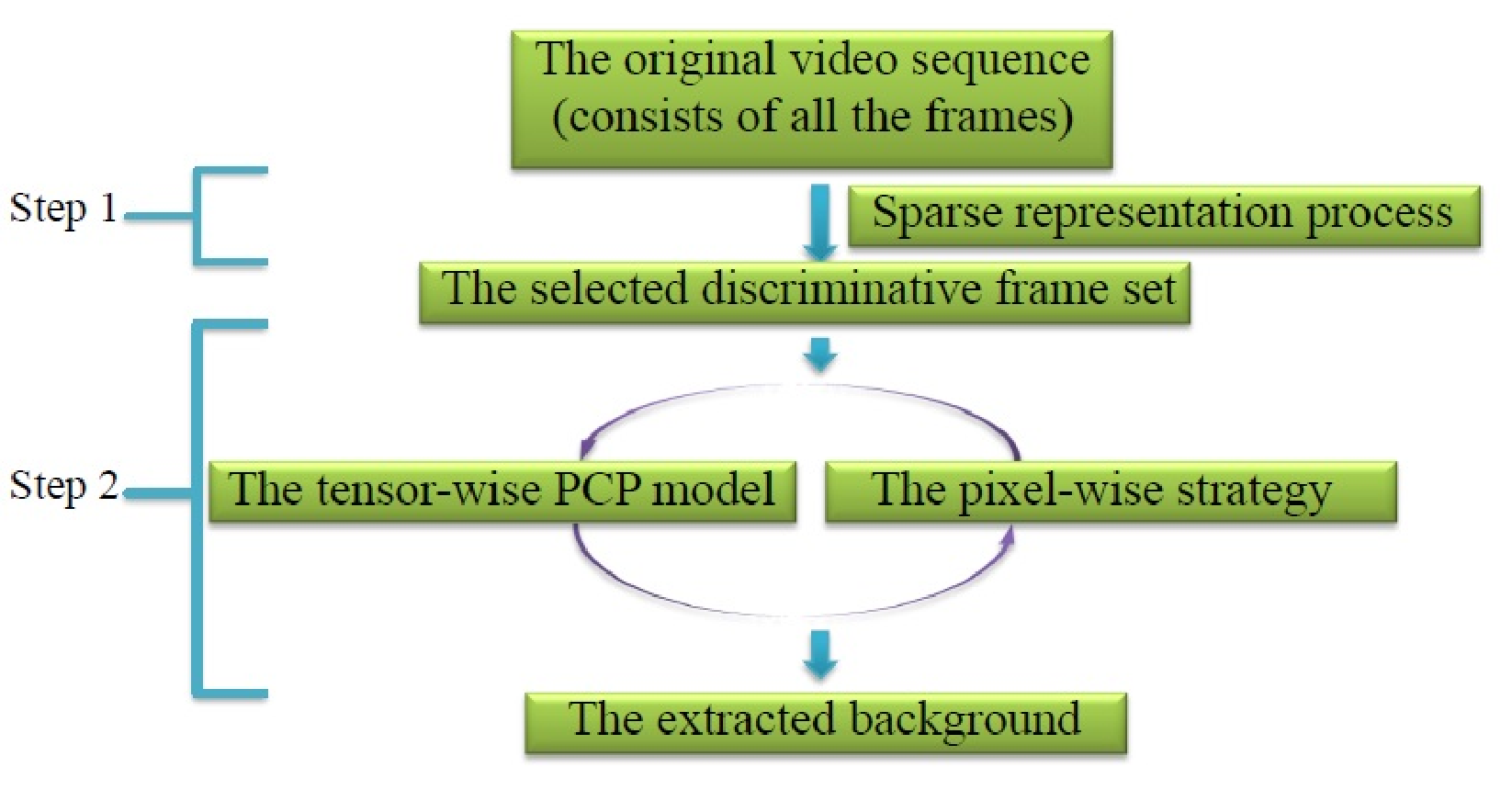}
\caption{\label{Fig.1}The framework of the Sparse Outlier Iterative Removal (SOIR) algorithm.}
\end{center}
\end{figure}

The main contributions of this work are summarized below:
\begin{itemize}
\item We utilize sparse representation to reduce the size of the video by exploring its ¡¯discriminative¡¯ frames. Instead of using all frames to model the background, we simply use the ¡¯discriminative¡¯ frames. In this way, our model can meet the demands of many practical background modeling problems in terms of speed and memory.
\item The cyclic iteration process is composed of a tensor-wise model and a pixel-wise strategy. In a general case, a tensor-wise process always considers the overall information, whereas a pixel-wise process pays more attention to particular information. Our algorithm achieves high accuracy by taking full advantages of both processes.
\item The tensor-wise model in the cyclic iteration is a PCP model, and is robust to general noises \cite{EXYJ2011}. Differing from previous works, the vectorized static background in our algorithm is explicit rank-1, instead of just being low-rank. To constrain this, we propose a new space $\mathcal{R}^{(4)}$ where the background actually lies. Owing to the rank-1 hypothesis, SVD is non-essential.

\end{itemize}

The remainder of this paper is organized as follows. Section \ref{sec2} introduces some Preliminary works. Section \ref{sec3} provides the formulation and convergence of the SOIR algorithm. Section \ref{sec4} presents the foreground detection method. Section \ref{sec5} describes the  experimental results. Finally, Section \ref{sec6} provides some concluding remarks regarding our research.

\section{Preliminary work}
\label{sec2}

The Principal Component Pursuit (PCP) model and tensor theory play key roles in our algorithm. Here, we introduce some basic works of both.

\subsection{Principal component pursuit}
\label{sec2.2}

Low-rank matrix recovery is the key problem in many practical tasks, including background modeling. A given data matrix $M$ is the superposition of a low-rank matrix $L$ and a sparse matrix $S$, i.e., $M=L+S$. Principal Component Analysis (PCA) is an effective way to solve this problem, but the brittleness of the original PCA model with respect to grossly corrupted observations jeopardizes its validity \cite{EXYJ2011}.

Cand\`{e}s et al. recently proved that one can recover matrix $L$ and the sparse matrix $S$ precisely under mild conditions \cite{EXYJ2011}. This model, known as Principal Component Pursuit (PCP), can be formulated as
\begin{equation}\label{eq2}
\begin{array}{l}
\mathop{\textrm{min}}\limits_{L , S}~\|L\|_\ast+\lambda\|S\|_{1}\\
 ~\textrm{s.t.} ~~~L+S=M,
\end{array}
\end{equation}
where $\lambda$ is a regularization parameter, $\|\cdot\|_\ast$ and $\|\cdot\|_{1}$ denote the nuclear norm (sum of singular values) and the $l_{1}$-norm (sum of the absolute values of the matrix elements), respectively.

Model (\ref{eq2}) was modified to solve the background modeling problem \cite{CTE2012,XCW2013}. All of the original works modeled the background using a low-rank matrix. In contrast, we consider the background as a rank-1 matrix.

\subsection{Tensors theory}
\label{sec2.3}

A tensor is a multidimensional array. More formally, a $N$-way or $N$-order tensor is an element of the tensor product of $N$ vector spaces, each of which has its own coordinate system\cite{BT2006,TB2009}. Intuitively, a vector is a 1-order tensor, while a matrix is a 2-order tensor. In this paper, in addition to the specific instructions, we denote vector by lowercase letter, e.g., $r$, and matrix by uppercase letter, e.g., $R$. In addition, a higher-order tensor is denoted in boldface type, e.g., $\mathbf{R}$. The space of all tensors is denoted by a swash font, e.g., $\mathcal{K}$. We denote the space of all $N$-order tensors by $\mathcal{R}_{N}$, $\mathcal{R}_{N}= \mathbb{R}^{K_{1} \times\cdots\times K_{N}}, N=1,2,3,4,\cdots$.

A tensor can be multiplied by a matrix, which is also known as the $n$-mode (matrix) product \cite{TB2009}. The $n$-mode product of tensor $\mathbf{X}\in\mathbb{R}^{K_{1} \times\cdots\times K_{N}}$ with matrix $U\in \mathbb{R}^{J\times K_{i}}$ is denoted by $\mathbf{X} \times_{i}U$ and is of size $K_{1}\times \cdots\times K_{i-1}\times J\times K_{i+1}\times \cdots \times K_{N}$. Element-wise, we have
\begin{equation}\label{eq7}
\begin{array}{l}
(\mathbf{X} \times_{i}U)_{k_{1}\cdots k_{i-1} j k_{i+1} \cdots k_{N}}=
\sum\limits_{k_{i}=1}^{K_{i}}\mathbf{X}_{k_{1}\cdots  k_{N}}U_{j k_{i}},
\end{array}
\end{equation}
where $k_{i}\in [1,\cdots,K_{i}]$($i= 1,\cdots,N$); $j\in [1,\cdots,J]$.

\section{Sparse outlier iterative removal algorithm}
\label{sec3}

In this section, we focus on modeling the background of a video sequence. We use $\mathbf{D}$ to denote a video, and assume that there are $N$ frames in $\mathbf{D}$. Each colorful frame is a 3-order tensor by nature, and the $j$-th frame is denoted by $\mathbf{I}_{j}\in \mathbb{R}^{m\times n\times 3}$. Then $\mathbf{D}= [\mathbf{I}_{1}, \cdots, \mathbf{I}_{N}] \in \mathbb{R}^{m\times n\times 3 \times N}$. In addition, we use $\mathbf{B}= [\mathbf{B}_{\mathbf{I}_{1}},\ldots, \mathbf{B}_{\mathbf{I}_{N}}]$ and $\mathbf{A}= [\mathbf{A}_{\mathbf{I}_{1}}, \ldots, \mathbf{A}_{\mathbf{I}_{N}}]$ to denote the background and foreground of video $\mathbf{D}$.

We first analyze the components of a video. In the video, the background is covered by the foreground objects. We denote the foreground region as $\Omega$, and the outside region as $\overline{\Omega}$. Let $\mathcal{P}_{\Omega}$ be an orthogonal projector onto the span of the tensors vanishing outside of $\Omega$. Then for an arbitrary frame $\mathbf{I}_{j}$, the $(x,y,z)$-th component of $\mathcal{P}_{\Omega}(\mathbf{I}_{j})$ is equal to $(\mathbf{I}_{j})_{xyz}$ if $(x, y, z)\in\Omega$, and is zero, otherwise. Thus, the video can be expressed as
\begin{equation}\label{eq8}
\mathbf{D}= \mathcal{P}_{\overline{\Omega}}(\mathbf{B})+ \mathcal{P}_{\Omega}(\mathbf{A}),
\end{equation}
where $\mathcal{P}_{\overline{\Omega}}(\mathbf{B})=[\mathcal{P}_{\overline{\Omega}} (\mathbf{B}_{\mathbf{I}_{1}}), \ldots ,\mathcal{P}_{\overline{\Omega}}(\mathbf{B}_{\mathbf{I}_{N}})]$ and $\mathcal{P}_{\Omega}(\mathbf{A}) = [\mathcal{P}_{\Omega}(\mathbf{A}_{\mathbf{I}_{1}}), \ldots, \mathcal{P}_{\Omega}(\mathbf{A}_{\mathbf{I}_{N}})]$. Actually, $\mathcal{P}_{\Omega}(\mathbf{A}_{\mathbf{I}_{i}})=\mathbf{A}_{\mathbf{I}_{i}}$ because $\Omega$ is simply the foreground region. The noise is also an aspect, i.e.,
\begin{equation}\label{eq9}
\mathbf{D}= \mathcal{P}_{\overline{\Omega}}(\mathbf{B})+ \mathbf{A}+\mathbf{E},
\end{equation}
where $\mathbf{E}$ is the noise. Equation (\ref{eq9}) shows the actual components of a video and is a strict constraint in our model.

\subsection{Discriminative exploration using sparse representation}
\label{sec3.1}

In most large-scale background modeling problems, the frames are highly redundant. Some of the frames already carry sufficient background information, and are more discriminative than other frames. In this section, we refine frame sequence $\mathbf{D}$ and obtain a new informative set $\widetilde{\mathbf{D}}$, which is composed of the selected discriminative frames.

We use sparse representation to explore the discriminative frames by solving the maximum linearly independent group of video frames. The sparse representation process, which is robust to noise \cite{JAASY2009}, is based on the video content. Once a frame is represented by other frames, its content is no longer discriminative. In a real-life video, the discriminative frames are those whose foreground objects are different in both position and appearance. Thus, we gain different background information from different discriminative frames. Once a sufficient amount of information is obtained,  we can model the background.

Now, we will introduce our sparse representation model. Foreground objects move continually in a video sequence. Two adjacent frames are usually approximately the same. Some frames can be represented through a linear combination of the remaining frames, and the other frames are usually repeated. In other words, a series of frames can represent all frames:
\begin{equation}\label{eq11}
\begin{array}{l}
\mathop{\textrm{min}}\limits_{C} ~\|\mathbf{D}-\mathbf{D} \times_{4} C\|_{F}^{2}+\lambda\|C\|_{1,2},\\
\end{array}
\end{equation}
where $\|\cdot\|_{F}$ is the Frobenius norm, which equals the square root of the sum of squares of the entries of the tensor. $C\in\mathbb{R}^{N\times N}$ is a coefficient matrix, and $\lambda$ is used to balance the two parts. In addition, $\|\cdot\|_{1,2}$ is the $l_{1,2}$-norm and is the sum of the $l_{2}$-norm of all rows in $C$ \cite{J2006}. We solve this model by converting it into an equivalent problem:
\begin{equation}\label{eq34}
\begin{array}{l}
\mathop{\textrm{min}}\limits_{W , C}~\|\mathbf{D}-\mathbf{D} \times_{4} W\|_{F}^{2}+\lambda\|C\|_{1,2},~~~
\textrm{s.t.} ~W=C.
\end{array}
\end{equation}
This problem is the standard augmented Lagrange formulation and can be solved using the Alternating Direction Multipliers (ADM) method \cite{XJ2013, ZRZ2011}.

The $j$-th row of $C$ records the coefficients of the $j$-th frame to represent other frames, and the $j$-th column of $C$ records the coefficients of other frames to represent the $j$-th frame. We can then deduce the role of each frame by observing the corresponding row in $C$. The frames whose coefficients are all zero are regarded as redundant, and the non-zero rows in $C$ correspond to discriminative frames. A new set $\widetilde{\mathbf{D}}$ is formed to contain all discriminative frames.

\begin{figure}[H]
\begin{center}
\includegraphics[width=0.45\textwidth]{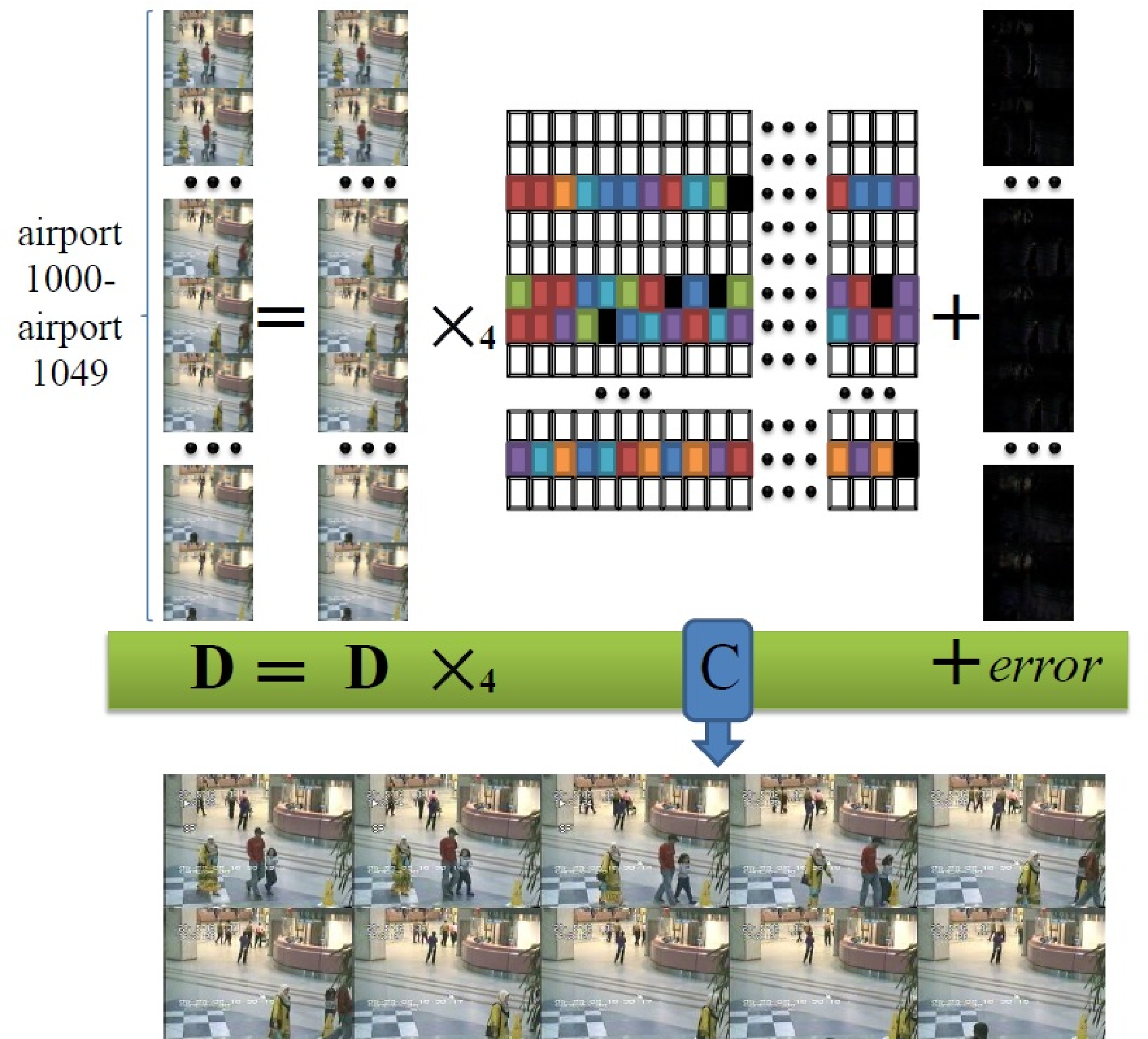}
\caption{\label{Fig.2}The sparse representation process of a discriminative exploration. The original frame set is composed of 50 frames (airport1000- airport1049 in the "hall" sequence from the I2R dataset), where the grid group indicates coefficient matrix $C$.}
\end{center}
\end{figure}
Fig. 2 illustrates our sparse representation process. A video sequence equals a sparse linear combination of itself plus errors. The color rows in the grid group indicate the non-zero rows in $C$. The corresponding frames of the nonzero rows in $C$ are selected, i.e., the ten frames in Fig. 2, which contain almost all background information of the 50 frames.

For most practical problems, the frames selected by Model (\ref{eq11}) are suitable for background modeling. However, some abnormal behaviors and serious noise pollution of the video may lead to more complicated relations among frames. For example, in Fig. 2, if the woman in yellow jumps from left to right, more frames will be considered discriminative. However, ten frames are sufficient for a background model. In this case, updating $\widetilde{\mathbf{D}}$ is essential. We do this based on coefficient matrix $C$, from which we can measure the similarity between two frames. If frame $\mathbf{I}_{a}$ resembles frame $\mathbf{I}_{b}$, the coefficient of frame $\mathbf{I}_{a}$ is close to 1 in representing frame $\mathbf{I}_{b}$; otherwise, it is far from 1. First, we find the frame that resembles all other frames in $\widetilde{\mathbf{D}}$ the most. This is the first reselected frame. Next, we choose the last similar frame of the first re-selected frame. Then, every time we choose a new frame, it is the last frame that all of the previously selected frames resemble. Eventually, we choose a new frame set in which the similarities among the members are low. This set is the updated $\widetilde{\mathbf{D}}$. The appropriate number of $\widetilde{\mathbf{D}}$ will be explored experimentally.

After the sparse representation of the video, we refine the original video $\mathbf{D}$ and form a new selected discriminative frame set $\widetilde{\mathbf{D}}$. Assume that there are $\widetilde{N}$ frames in $\widetilde{\mathbf{D}}$ : $ \widetilde{\mathbf{D}}\in \mathbb{R}^{m\times n\times 3 \times \widetilde{N}}$, where $ \widetilde{N}\ll N$.

\subsection{Background extraction using cyclic iteration process}
\label{sec3.2}

\begin{figure}[H]
\begin{center}
\includegraphics[width=0.43\textwidth]{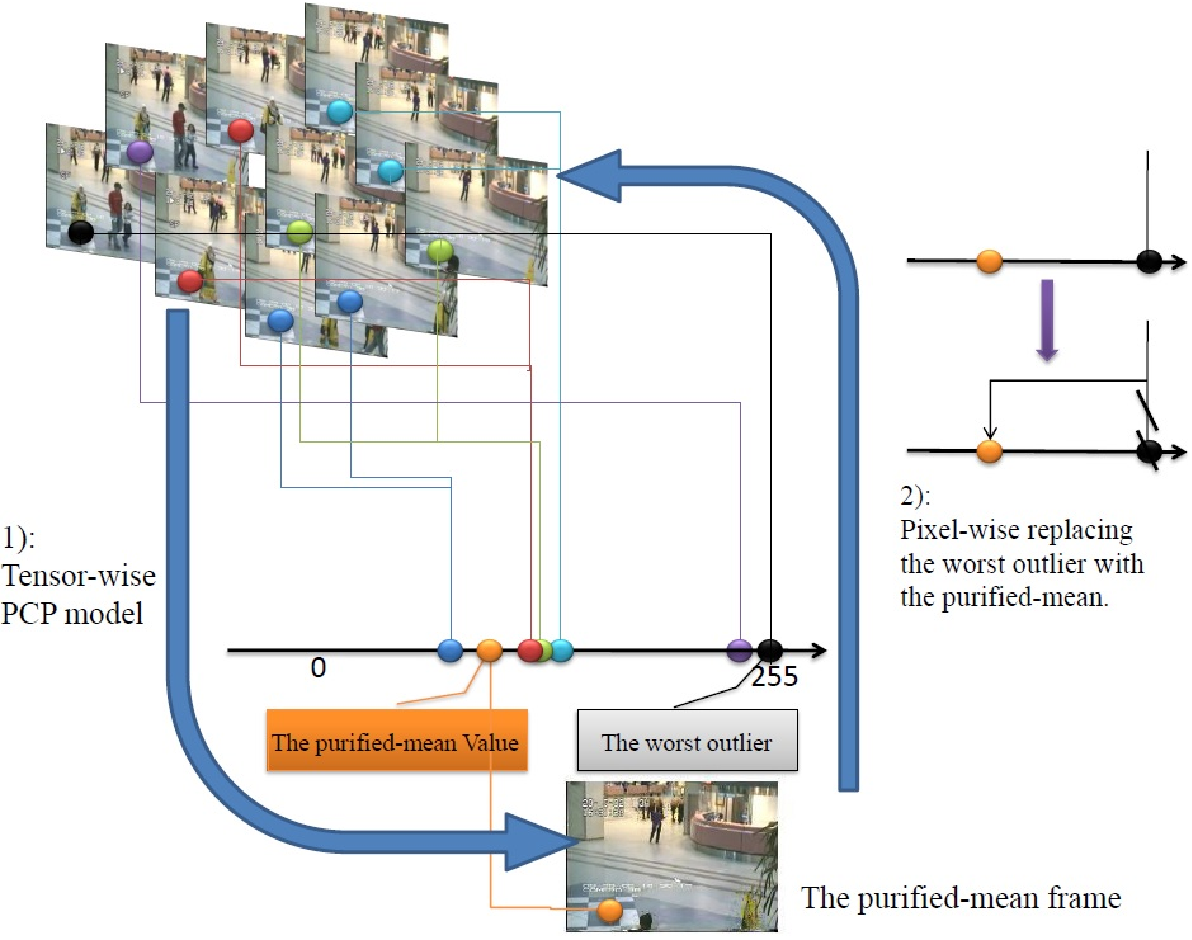}
\caption{\label{Fig.3}The cyclic iteration process in an iteration.}
\end{center}
\end{figure}
In this section, we describe the design of a background extraction using a cyclic iteration (the outer loop in SOIR). This process is shown in Fig. 3. In each iteration, we use a PCP model to solve the purified-mean frame from the selected discriminative frame set $\widetilde{\mathbf{D}}$, and in a pixel-wise outlier removal strategy, we use the purified-mean frame to ameliorate the selected discriminative frame set $\widetilde{\mathbf{D}}$ conversely. The iteration will continue until the purified-mean frames converge to a fixed frame, which is the background.

\subsubsection{Tensor-wise PCP model}
\label{sec3.2.2}

The tensor model is used to calculate the purified-mean of the selected discriminative frames. It is the mean of the frames initially, but moves slightly away during the denoising process. Following the idea of ADM, we solve this tensor model through an iterative approximation (the inner loop in SOIR). The solution is limited to a $\mathcal{R}^{(4)}$ space.

The purified-mean frame, denoted by $\mathbf{B}^{\ast}:\mathbf{B}^{\ast}\in R^{m \times n \times 3}$,  is the optimal background of the current frame set $\widetilde{\mathbf{D}}$. The real backgrounds of different frames are the same, i.e., $\mathbf{B}_{\mathbf{I}_{i}}= \mathbf{B}_{\mathbf{I}_{j}}=\mathbf{B}^{\ast}, \forall i\neq j$, or the following:
\begin{equation}\label{eq22}
\widetilde{\mathbf{B}}=\mathbf{B}^{\ast} \times_{4}\widetilde{Z},
\end{equation}
where $\widetilde{Z}=(1,1,\ldots,1)^{\textrm{T}}\in \mathbb{R}^{\widetilde{N}\times 1}$ is a 1-order matrix. $\mathbf{B}^{\ast}$ is regarded as a 4-order tensor, i.e., $\mathbf{B}^{\ast}\in \mathbb{R}^{m\times n\times 3 \times 1}$.

Constraint (\ref{eq22}) in fact insists that the background is rank-1. Just like the vectorizing process in previous works  \cite{CTE2012,EGR2012,CXW2011}, we transform the frames into gray-scales and combine mode-1 and mode-2 of each frame into a single mode. This means the operator $vectorize(\cdot)$ reduces the dimension of high-dimensional data. After the vectorizing process, a video tensor is transformed into a matrix, and a frame tensor is transformed into a vector. The vectorized formulation of Equation (\ref{eq22}) is then
\begin{equation}\label{eq14}
vectorize(\widetilde{\mathbf{B}})=vectorize(\mathbf{B}^{\ast}) \times_{2}\widetilde{Z},
\end{equation}
where $vectorize(\widetilde{\mathbf{B}})\in R^{p \times \widetilde{N}}$ is a matrix ,$vectorize(\mathbf{B}^{\ast})\in R^{p}$ is a vector and the 2-mode product ($\times_{2}$) is the outer product of the vectors. Thus, the equation reduces to the standard definition of rank-1.

To solve Constraint (\ref{eq22}), we consider a subspace of $\mathcal{R}_{4}$.  We denote this subspace by $\mathcal{R}^{(4)}$. All tensors in $\mathcal{R}^{(4)}$ are 4-order tensors, and for each tensor $\mathbf{X}$ in this space, element-wise, we have $\mathbf{X}_{ijka}= \mathbf{X}_{ijkb},\forall a\neq b$. Thus, $\mathbf{B}^{\ast}$ must lie within this space. $\mathcal{R}^{(4)}$ is convex, and it is therefore easy to solve (\ref{eq22}).
\begin{lemma}
Given tensor $\mathbf{B}$, the solution to the problem
\begin{equation}\label{eq26}
\mathop{\textrm{min}}\limits_{\mathbf{B}^{\ast}} \|\widetilde{\mathbf{B}}-\mathbf{B}^{\ast} \times_{4}\widetilde{Z}\|_{F}^{2}
\end{equation}
is $\mathbf{B}^{\ast}=(\sum_{l=1}^{\widetilde{N}}\widetilde{\mathbf{B}}_{\widetilde{\mathbf{I}}_{l}})/\widetilde{N}$.
\end{lemma}
\begin{IEEEproof} $\|\widetilde{\mathbf{B}}-\mathbf{B}^{\ast} \times_{4}\widetilde{Z}\|_{F}^{2} = \sum_{l=1}^{\widetilde{N}}\|\widetilde{\mathbf{B}}_{\widetilde{\mathbf{I}}_{l}}- \mathbf{B}^{\ast}\|_{F}^{2} = \widetilde{N} \|\mathbf{B}^{\ast}- (\sum_{l=1}^{\widetilde{N}}\widetilde{\mathbf{B}}_{\widetilde{\mathbf{I}}_{l}})/\widetilde{N} \|_{F}^{2} + const$, where $const$ and $\widetilde{N}$ indicate constants. This completes the proof.
\end{IEEEproof}
To model the background, we want the static video content. We therefore minimize the changing part to group more information into the background. In addition, we take strict Constraints (\ref{eq9}) and (\ref{eq22}) into account and give our model:
\begin{equation}\label{eq10}
\begin{array}{l}
\mathop{\textrm{min}}\limits_{\widetilde{\mathbf{B}},\widetilde{\mathbf{A}}, \widetilde{\mathbf{E}}}~\|\widetilde{\mathbf{A}}+\widetilde{\mathbf{E}}- \mathcal{P}_{\Omega}(\widetilde{\mathbf{B}})\|_{1}\\
  \textrm{s.t.}
  ~\widetilde{\mathbf{D}}= \mathcal{P}_{\overline{\Omega}}(\widetilde{\mathbf{B}})+ \widetilde{\mathbf{A}}+\widetilde{\mathbf{E}}\\
  ~~~~~\widetilde{\mathbf{B}}=\mathbf{B}^{\ast} \times_{4}\widetilde{Z}.
\end{array}
\end{equation}
In the foreground region, we minimize the number of nonzero elements in $\widetilde{\mathbf{A}}+\widetilde{\mathbf{E}}- \widetilde{\mathbf{B}}$; otherwise, such pixels can be considered as background if $\widetilde{\mathbf{A}}+\widetilde{\mathbf{E}}- \widetilde{\mathbf{B}}=0$. Outside the foreground region, we minimize the noise $\widetilde{\mathbf{E}}$. Benefitting from the definition of the $l_{1}$-norm, we arrange the two regions into a single formula, i.e., the objective function in Model (\ref{eq10}).

Model (\ref{eq10}) is a PCP model. To solve this model, we first arrange it. The constraint $\widetilde{\mathbf{B}}=\mathbf{B}^{\ast} \times_{4}\widetilde{Z}$ can not be transformed into a single variable linear equation. We therefore use it as a correction term. In addition, we denote all non-background parts by $\widetilde{\mathbf{S}}: \widetilde{\mathbf{S}} = \widetilde{\mathbf{A}}+ \widetilde{\mathbf{E}}- \mathcal{P}_{\Omega}(\widetilde{\mathbf{B}})$. We then obtain the following:
\begin{equation}\label{eq13}
\begin{array}{l}
\mathop{\textrm{min}}\limits_{\widetilde{\mathbf{B}},\widetilde{\mathbf{S}}}~\|\widetilde{\mathbf{S}} \|_{1},
  ~~\textrm{s.t.} ~~\widetilde{\mathbf{D}}= \widetilde{\mathbf{B}}+ \widetilde{\mathbf{S}}.
\end{array}
\end{equation}
Model (\ref{eq13}) can be solved by the iteration \cite{XJ2013} (the inner loop):
\begin{equation}\label{eq19}
\left\{
\begin{array}{l}
\widetilde{\mathbf{S}}^{k+1}=\mathcal{T}_{\frac{1}{\mu}}(\widetilde{\mathbf{D}}+\frac{\widetilde{\mathbf{\Lambda}}^{k}}{\mu}-\widetilde{\mathbf{B}}^{k})\\
\widetilde{\mathbf{B}}^{k+1}=\widetilde{\mathbf{D}}-\widetilde{\mathbf{S}}^{k+1}\\
\widetilde{\mathbf{\Lambda}}^{k+1}=\widetilde{\mathbf{\Lambda}}^{k}-\mu(\widetilde{\mathbf{D}}-\widetilde{\mathbf{B}}^{k+1}-\widetilde{\mathbf{S}}^{k+1}),
\end{array}
\right.
\end{equation}
where $\mu>0$ is a step-length parameter and $\widetilde{\mathbf{\Lambda}}$ is the Lagrange multiplier. In addition, $\mathcal{T}_{\frac{1}{\mu}}(\cdot)$ is a soft-threshold operator \cite{XJ2013, ZRZ2011}. For an arbitrary tensor $\mathbf{X}\in\mathbb{R}^{K_{1} \times\cdots\times K_{N}}$, element-wise, the operator is given by
\begin{equation}\label{eq21}
\mathcal{T}_{\frac{1}{\mu}}(\mathbf{X})_{{k_{1} \cdots k_{N}}}=\textrm{max}(\mathbf{X}_{{k_{1} \cdots k_{N}}}- \frac{1}{\mu},0)+ \textrm{min}(\mathbf{X}_{{k_{1} \cdots k_{N}}}+ \frac{1}{\mu},0).
\end{equation}

Here, we consider the  correction term. The tensor $\widetilde{\mathbf{B}}$ must lie in the $\mathcal{R}^{(4)}$ space. Once we obtian a new $\widetilde{\mathbf{B}}$ in (\ref{eq19}), we project it into the $\mathcal{R}^{(4)}$ space and use its vertical projection to replace itself. Thus, the updated formula of $\widetilde{\mathbf{B}}^{k+1}$ in (\ref{eq19}) is replaced by
\begin{equation}\label{eq24}
\widetilde{\mathbf{B}}^{k+1}= \mathbf{B}^{\ast k+1} ,~(\mathbf{B}^{\ast k+1}:\widetilde{\mathbf{D}}-\widetilde{\mathbf{S}}^{k+1}= \mathbf{B}^{\ast k+1} \times_{4}\widetilde{Z}). \\
\end{equation}

The result of the inner loop is the optimal background $\mathbf{B}^{\ast}$ of the current frame set $\widetilde{\mathbf{D}}$.

\subsubsection{Pixel-wise strategy}
\label{sec3.2.1}
Once we obtain the purified-mean frame $\mathbf{B}^{\ast}$ of the current frame set $\widetilde{\mathbf{D}}$, we use $\mathbf{B}^{\ast}$ to renew the current frame set $\widetilde{\mathbf{D}}$ conversely. This is a pixel-wise method, called outlier removal strategy. We repeat this strategy for all pixels in each frame. As an example, we then take the pixel ($x,y,z$) ($x= 1,\cdots,m;y=1,\cdots,n;z=1,2,3$).

In Fig. 3, we have labeled pixel ($x,y,z$) in all frames with solid color points. Different colors indicate different pixel values. When we locate these values in the axis, we find that most of the values gather into a cluster, and others do not. The outliers are the pixel values of the non-background pixels. The purified-mean value and worst outlier are also marked in the figure. The worst outlier is the value that is farthest away from the purified-mean value.

The ground truth of the background is inside the cluster. The purified-mean value is much closer to the ground truth than the worst outlier. Thus, the extraction performance will be improved if we use the purified-mean value to replace the worst outlier. As shown in Fig. 3, once we extract a purified-mean frame $\mathbf{B}^{\ast}$, we continue replacing the worst outlier with the purified-mean value for each pixel. Here, replacing the worst outlier means deleting the worst value and returning the purified-mean value to the frame. Thus, frame set $\widetilde{\mathbf{D}}$ is renewed after the replacement is conducted for all pixels.

\subsection{Algorithm formulation}
\label{sec3.3}

The methods in Sections \ref{sec3.1} and \ref{sec3.2} combine to form our Sparse Outlier Iterative Removal (SOIR) algorithm. In this algorithm, the sparse representation model is an important aspect. It returns the selected discriminative frames that carry sufficient background information. The cyclic iteration process is the main aspect, which extracts the background of the video. The convergence condition of the algorithm is $\|\widetilde{\mathbf{B}}^{k+1}-\widetilde{\mathbf{B}}^{k}\|/\|\widetilde{\mathbf{B}}^{k}\|\leq1e$-$3$.

\begin{table}[H]\label{alg1}
\centering
\begin{tabular}{l}
\hline $\textbf{Algorithm ~1:}~~\textrm{SOIR Algorithm.}$\\
\hline
$\textbf{Input:}~\mathbf{D}\in \mathbb{R}^{m\times n\times 3 \times N}$.\\
$\textbf{output:}~\mathbf{B}^{\ast}$.\\
1:$\textbf{sparse representation:}$\\
~~~~~$ get ~\widetilde{\mathbf{D}}$, ~$by~solving$:\\
~~~~~~~~~~$\mathop{\textrm{min}}\limits_{C}~\|\mathbf{D}-\mathbf{D} \times_{4} C\|_{F}^{2}+\lambda\|C\|_{1,2}$. \\
2:$\textbf{cyclic iteration:}$\\
~~~$\textbf{while~not~converged}~\textbf{do} (outer ~loop):$\\
~~~(1):~$update~~~\mathbf{B}^{\ast}, ~~~by$:\\
~~~~~$\textbf{while~not~converged}~\textbf{do} (inner ~loop):$\\
~~~~~~~~$\widetilde{\mathbf{S}}^{k+1}=\mathcal{T}_{\frac{1}{\mu}}(\widetilde{\mathbf{D}}+\frac{\widetilde{\mathbf{\Lambda}}^{k}}{\mu}-\widetilde{\mathbf{B}}^{k})$;\\
~~~~~~~~$\widetilde{\mathbf{B}}^{k+1}=\mathbf{B}^{\ast k+1}$, $where$ \\
~~~~~~~~~~~~~$\mathbf{B}^{\ast k+1}= (\sum_{l=1}^{\widetilde{N}}(\widetilde{\mathbf{D}}_{\widetilde{\mathbf{I}}_{l}}-\widetilde{\mathbf{S}}^{k+1}_{\widetilde{\mathbf{I}}_{l}}))/\widetilde{N}$;\\
~~~~~~~~$\widetilde{\mathbf{\Lambda}}^{k+1}=\widetilde{\mathbf{\Lambda}}^{k}-\mu(\widetilde{\mathbf{D}}-\widetilde{\mathbf{B}}^{k+1}-\widetilde{\mathbf{S}}^{k+1})$.\\
~~~~~$\textbf{end~while}$.\\
~~~(2):~$ update~~~ \widetilde{\mathbf{D}}~ (\widetilde{\mathbf{D}}=[\widetilde{\mathbf{I}_{1}}, \cdots, \widetilde{\mathbf{I}_{\widetilde{N}}}]),~~~ by$:\\
~~~~~~~~$\textbf{For}~ pixel~(x,y,z)$\\
~~~~~~~~~~~~~~$N^{\ast}= \textrm{argmax}_{\widetilde{i}} |(\widetilde{\mathbf{I}_{\widetilde{i}}})_{xyz}- {\mathbf{B}}_{xyz}^{\ast}|$,\\
~~~~~~~~~~~~~~$then ~~(\widetilde{\mathbf{I}_{N^{\ast}}})_{xyz}= {\mathbf{B}}_{xyz}^{\ast}$.\\
~~~~~~~~$\textbf{End}$\\
~~~$\textbf{end~while}$.\\
\hline
\end{tabular}
\end{table}

\subsection{Convergence analysis}
\label{sec3.4}

We will now describe the convergence of SOIR. The convergence of a sparse representation model has been well studied \cite{JFJG2009}. Thus, we focus on the convergence of the cyclic iteration process.

For an arbitrary pixel ($x,y,z$), there are $\widetilde{N}$ pixel values in the selected set $\widetilde{\mathbf{D}}$. In the $i$-th outer loop iteration, the minimum and maximum of the $\widetilde{N}$ values are recorded as $a_{i}$ and $b_{i}$. $\mathbf{B}_{xyz}^{\ast i}$ is the purified-mean of the $\widetilde{N}$ values in last iteration, and thus $\mathbf{B}_{xyz}^{\ast i}\in [a_{i-1}, b_{i-1}]$. If $\lim_{i\rightarrow \infty} (b_{i}-a_{i})=0$, the purified-mean $\mathbf{B}_{xyz}^{\ast i}$ will converge. This can be inferred from the Nested Intervals Theorem \cite{RW2012}. To complete the convergence, we prove the following lemma.

\begin{lemma}
In the SOIR algorithm, for an arbitrary pixel ($x,y,z$), the minimum and maximum of the $\widetilde{N}$ pixel values in the $i$-th iteration are recorded as $a_{i}$ and $b_{i}$. we then have $\lim_{i\rightarrow \infty} (b_{i}-a_{i})=0$.
\end{lemma}
\begin{IEEEproof} First, the purified-mean value $ \widetilde{\mathbf{B}}_{xyz}^{i+1}$ is inside a subinterval of the interval $[a_{i}, b_{i}]$, because the value is simply around the mean of all the $\widetilde{N}$ values in last iteration. Otherwise, if the mean of the $\widetilde{N}$ values is close to either their minimum or maximum, we can conclude that the $\widetilde{N}$ values are already close to each other\cite{RW2012}.

Second, after $\widetilde{N}$ iterations, we record the minimum and the maximum of all the $\widetilde{N}$ purified-mean values ($ \widetilde{\mathbf{B}}_{xyz}^{i}, i=1,2, \ldots, \widetilde{N}$) as $c$ and $d$. Then, we have $[a_{i+\widetilde{N}},b_{i+\widetilde{N}}]\subseteq [c, d]$, because the worst outlier is replaced by the purified-mean value in each iteration, and all of the $\widetilde{N}$ purified-mean values are in the interval $[c,d]$. The subinterval $[c,d]$ is shorter than the interval $[a_{i}, b_{i}]$. In other words, there is a constant ratio $\gamma: \gamma<1$, subject to $(b_{i+\widetilde{N}}-a_{i+\widetilde{N}})\leq \gamma(b_{i}-a_{i})$.

Finally, we assume that the original minimum and maximum of the $\widetilde{N}$ values are $a$ and $b$, respectively. Then $(b_{1+\widetilde{N}}-a_{1+\widetilde{N}}) \leq  \gamma(b-a) $, $(b_{1+2\widetilde{N}}-a_{1+2\widetilde{N}}) \leq  \gamma^{2} (b-a)$, $\cdots$, $(b_{1+n\widetilde{N}}-a_{1+n\widetilde{N}}) \leq  \gamma^{n} (b-a)$, $\cdots$. We know that $(b-a)$ is a constant, and $\gamma^{n}$ is close to zero when $n$ is large. We then have $\lim_{i\rightarrow \infty} (b_{i}-a_{i})=0$, which completes the proof.
\end{IEEEproof}

$~$

We have to point out that the derived solution may not be the ground truth,  and is influenced by the property of the video. The experiments in Section \ref{sec5} will show that the solution is pretty close to the ground truth if the video quality is not too poor.

\section{Foreground region detection}
\label{sec4}
Having computed the background tensor, as described in Section \ref{sec3}, we then detect the foreground region of the video.

\label{sec4.1}
\begin{figure*}[htb]
\centering
\includegraphics[width=0.9\textwidth]{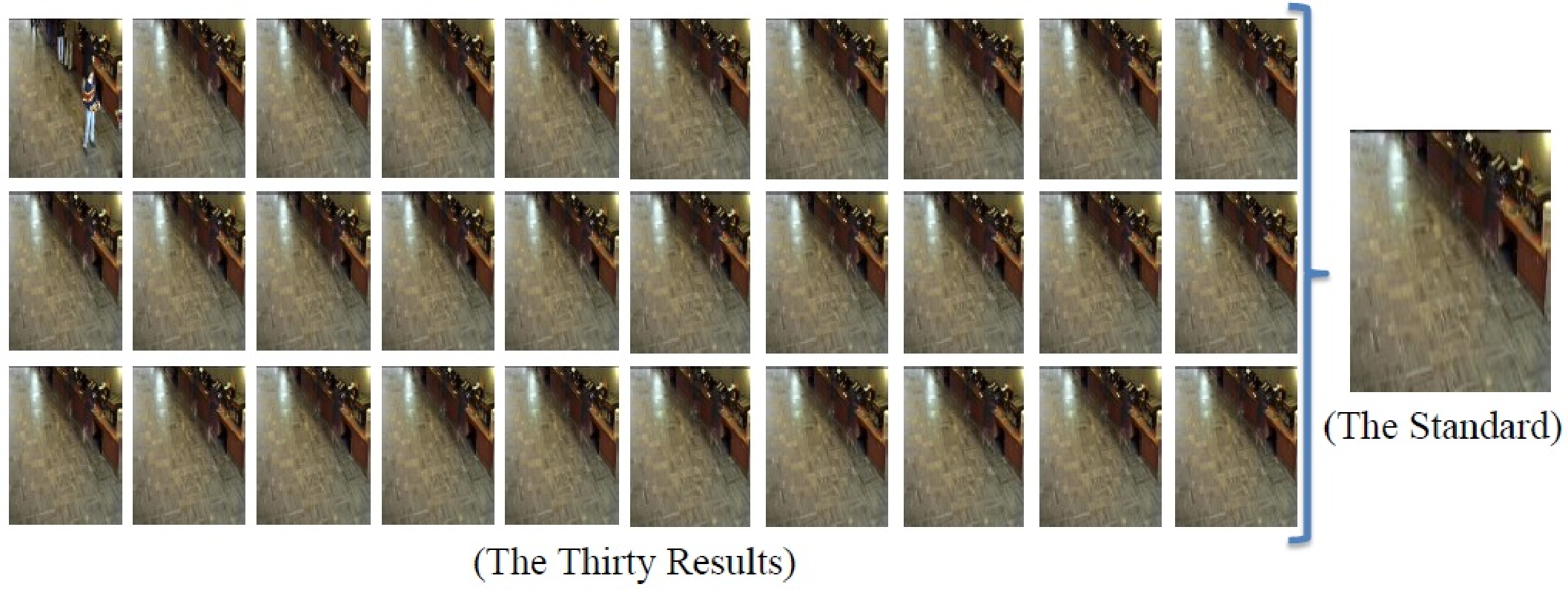}
\caption{\label{Fig.4}The background extracting results with the size of the selected set varying from 1 to 30. The standard background is extracted when using all the 300 frames as the selected set.}
\end{figure*}

\begin{figure*}[htb]
\begin{center}
\subfigure[]{\includegraphics[width=0.33\textwidth]{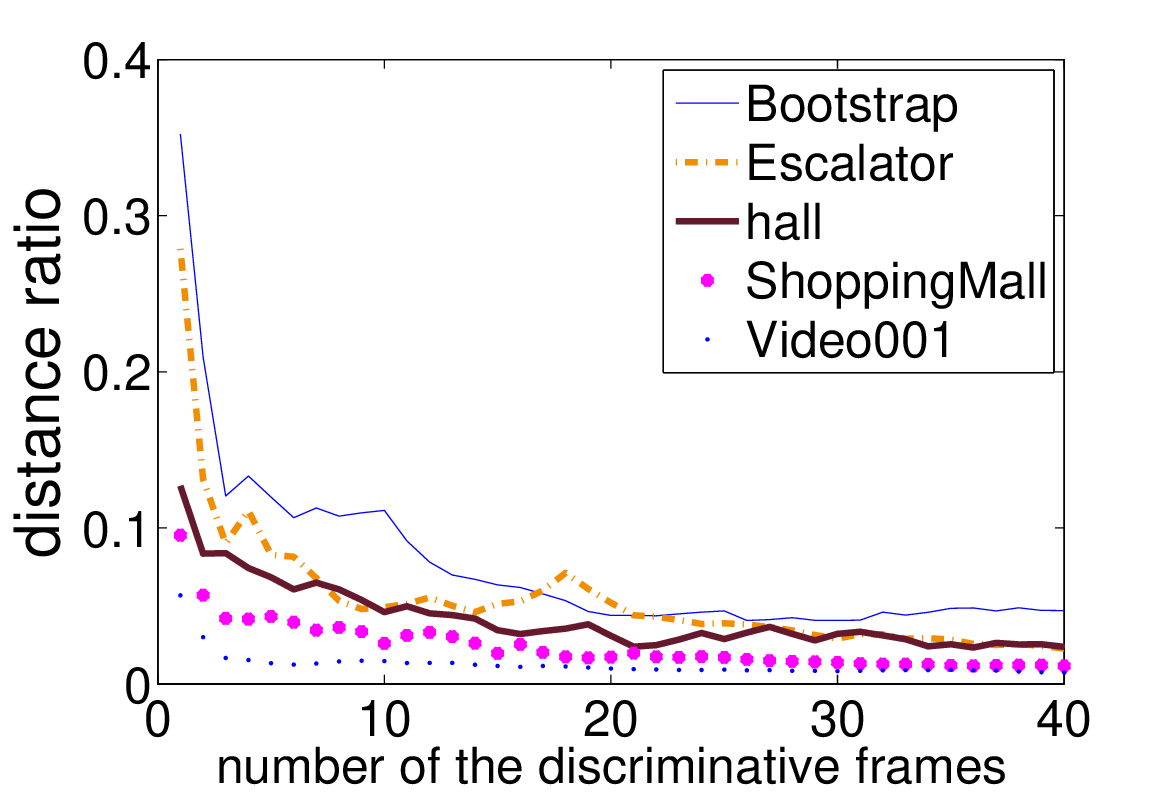}}
\subfigure[]{\includegraphics[width=0.33\textwidth]{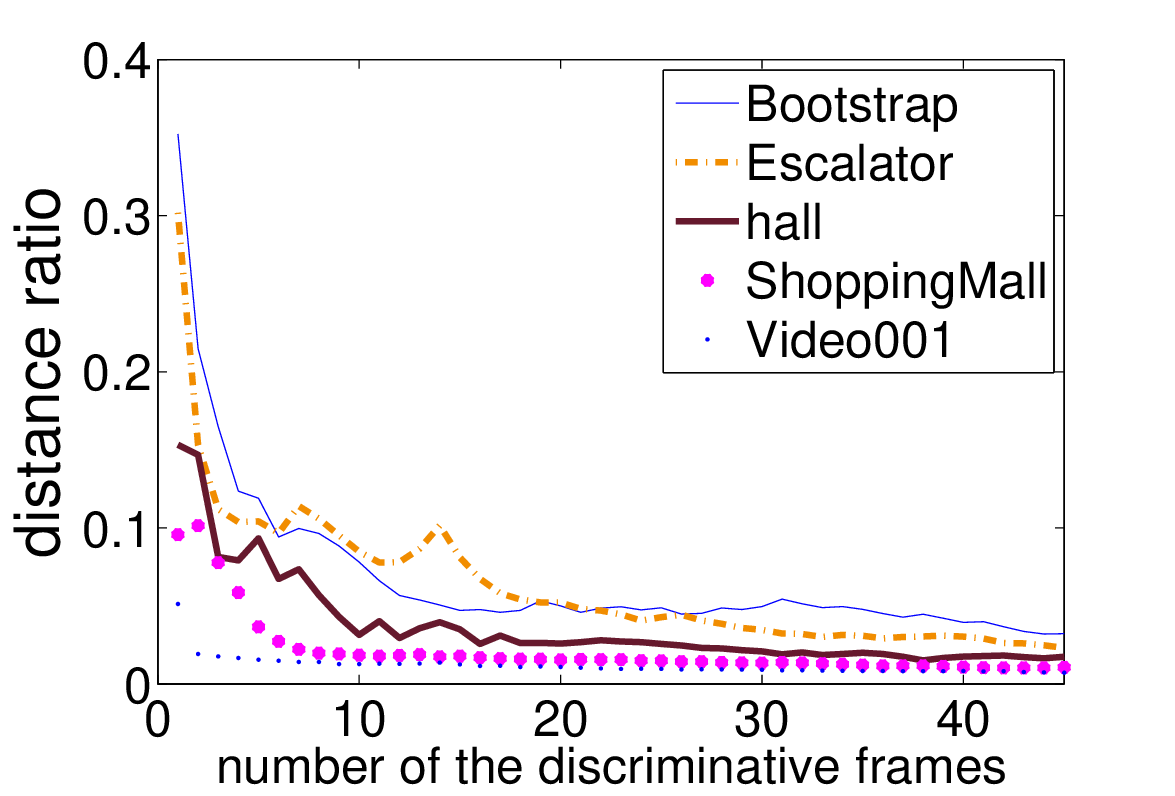}}
\subfigure[]{\includegraphics[width=0.33\textwidth]{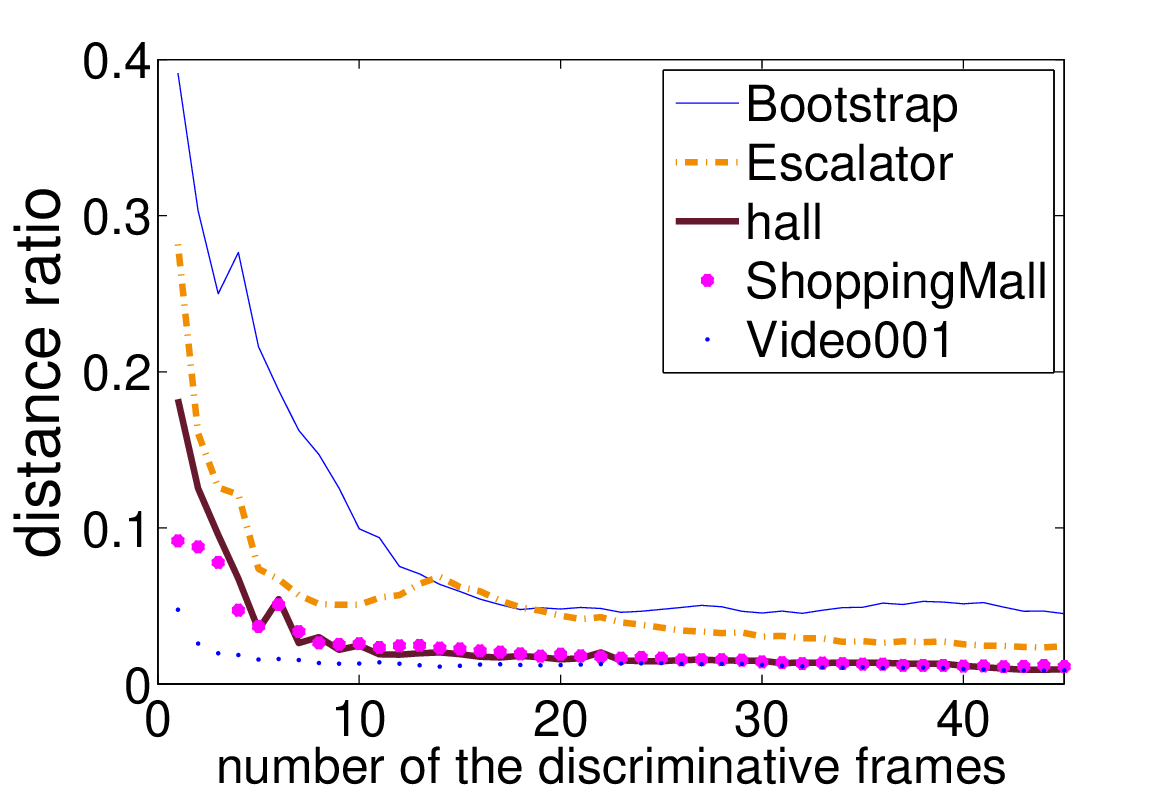}}
\subfigure[]{\includegraphics[width=0.33\textwidth]{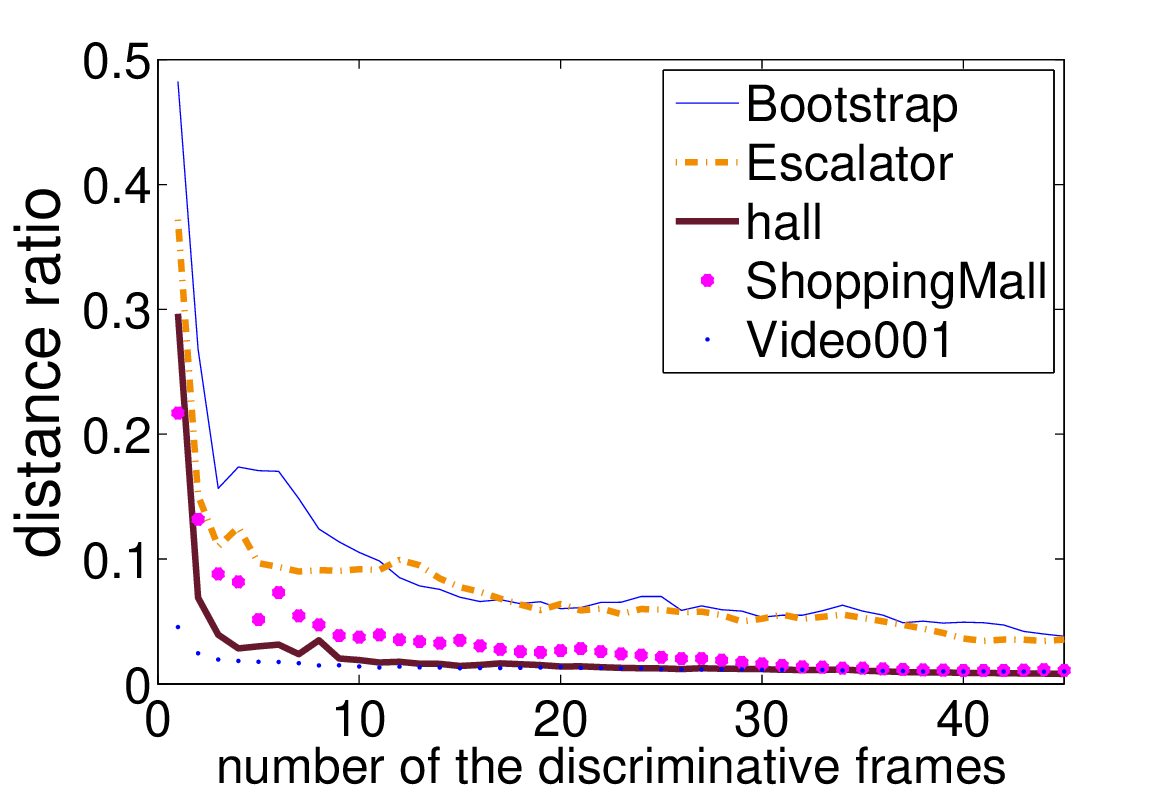}}
\caption{\label{Fig.6}The relationship between the number and the performance (2):(a) the original frame number is 450; (b) the original frame number is 600; (c) the original frame number is 750; (d) the original frame number is 900.}
\end{center}
\end{figure*}

Background subtraction is a common method for detecting the foreground region. We denote the result of the subtraction as $\mathbf{F}_{\mathbf{I}_{k}}$: $\mathbf{F}_{\mathbf{I}_{k}}  = \mathbf{I}_{k}-\mathbf{B}_{\mathbf{I}_{k}}$, where
$\mathbf{I}_{k}= \mathcal{P}_{\overline{\Omega}}(\mathbf{B}_{\mathbf{I}_{k}})+ \mathbf{A}_{\mathbf{I}_{k}}+ \mathbf{E}_{\mathbf{I}_{k}}$ and $\mathbf{B}_{\mathbf{I}_{k}}= \mathbf{B}^{\ast}$.  We found that the residual background only exists in the foreground region, and outside this region nothing but noise exists, i.e.,
\begin{equation}\label{eq29}
\begin{array}{l}
\mathbf{F}_{\mathbf{I}_{k}}
=\left\{
\begin{array}{rcl}
\mathbf{A}_{\mathbf{I}_{k}}+ \mathbf{E}_{\mathbf{I}_{k}}- \mathbf{B}_{\mathbf{I}_{k}} &, & \textrm{inside the region $\Omega$;}\\
\mathbf{E}_{\mathbf{I}_{k}}&,& \textrm{outside the region $\Omega$}.
\end{array}
\right.
\end{array}
\end{equation}

From Expression (\ref{eq29}), we can conclude that the background subtraction method works depending on the properties of $\mathbf{A}_{\mathbf{I}_{k}}- \mathbf{B}_{\mathbf{I}_{k}}$ and $\mathbf{E}_{\mathbf{I}_{k}}$, and the relationship between them. If the distribution of $\mathbf{A}_{\mathbf{I}_{k}}-\mathbf{B}_{\mathbf{I}_{k}}$ is different from that of $\mathbf{E}_{\mathbf{I}_{k}}$, the background subtraction will be an impactful way to detect the foreground.

We next explore the foreground region $\Omega$ for an arbitrarily given image $\mathbf{I}_{k}$ from the original frame set $\mathbf{D}$. To simplify this problem, we transform the color frame into gray one ($I_{k}$). We model the region using Markov Random Field, following previous works \cite{XCW2013,S2009, CH2010}.

First, we set up a matrix $O$ to represent the foreground region $\Omega$:
\begin{equation}\label{eq30}
\begin{array}{l}
O_{ij}
=\left\{
\begin{array}{rcl}
1 &, & (\mathbf{A}_{{I}_{k}})_{ij} \neq 0\\
0&,& (\mathbf{A}_{{I}_{k}})_{ij} = 0.
\end{array}
\right.
\end{array}
\end{equation}
The energy of $\Omega$ can then be obtained using the Ising model\cite{S2009}:
\begin{equation}\label{eq31}
\sum_{i,j}\lambda_{a}*O_{ij}+\sum_{i,j,x,y:|i-x|+|j-y|\leq1}\lambda_{b}*|O_{ij}-O_{xy}|,
\end{equation}
where $\lambda_{a}$ and $\lambda_{b}$ are two positive parameters, that penalize $O_{ij}=1$ and $|O_{ij}-O_{xy}|=1$, respectively.

Clearly, if we simply minimize the energy of the foreground region $\Omega$, it will converge to an empty set, i.e., $O=0$. To avoid this, an important component of the objective function is $\mathcal{P}_{\overline{\Omega}}(\mathbf{F}_{I_{k}})$. In addition, the non-zero elements outside the foreground region should also be minimized. Thus, we have the following foreground detection model:
\begin{equation}\label{eq32}
\begin{array}{l}
\mathop{\textrm{min}}\limits_{O_{ij}}\sum\limits_{O_{ij}=0}\frac{1}{2}((I_{k})_{ij}-(\mathbf{B}_{I_{k}})_{ij})^2+\sum\limits_{i,j}\lambda_{a}*O_{ij}\\
~~~~~~~~~~~~+\sum\limits_{i,j,x,y:|i-x|+|j-y|\leq1}\lambda_{b}*|O_{ij}-O_{xy}|.
\end{array}
\end{equation}
Model (\ref{eq32}) can be rearranged as
\begin{equation}\label{eq33}
\begin{array}{l}
\mathop{\textrm{min}}\limits_{O_{ij}}\sum\limits_{ij} (\lambda_{a}-\frac{1}{2}((I_{k})_{ij}-(\mathbf{B}_{I_{k}})_{ij})^2)*O_{ij}\\
~~~~~~~~~~~~~~~~+\lambda_{b}*\|A \times_{3} O\|_{1},
\end{array}
\end{equation}
where $A$ is a projection tensor. The constant part of Model (\ref{eq32}) is omitted because it is insignificant in an optimization problem. Model (\ref{eq33}) is the standard form of a first-order Markov Random Field, and can be solved exactly using graph cuts \cite{YOR2001}.

\section{Experimental analysis}
\label{sec5}

In this section, we describe the performance evaluation of our Sparse Outlier Iterative Removal (SOIR) algorithm. To evaluate the algorithm, we will explore the appropriate number of discriminative frames and test the performance and time consumption of the algorithm. The experiments are conducted on real sequences from public datasets such as the I2R \cite{LWIQ2004}, flowerwall \cite{KJBB1999}, SABS \cite{SBG2013}, and BMC datasets \cite{ATAL2012}. In addition, some public video sequences from the Internet are also included in our experiments. All experiments are conducted and timed in Matlab R2010a on a PC with a 3.20GHz Intel(R) Core(TM) CPU and 4GB of RAM.

\subsection{Number of discriminative frames}

A major aspect of this paper is utilizing sparse representation to reduce the size of the video. In this section, we explore the appropriate number of discriminative frames.

\begin{figure}[H]
\begin{center}
\subfigure[]{\includegraphics[width=0.33\textwidth]{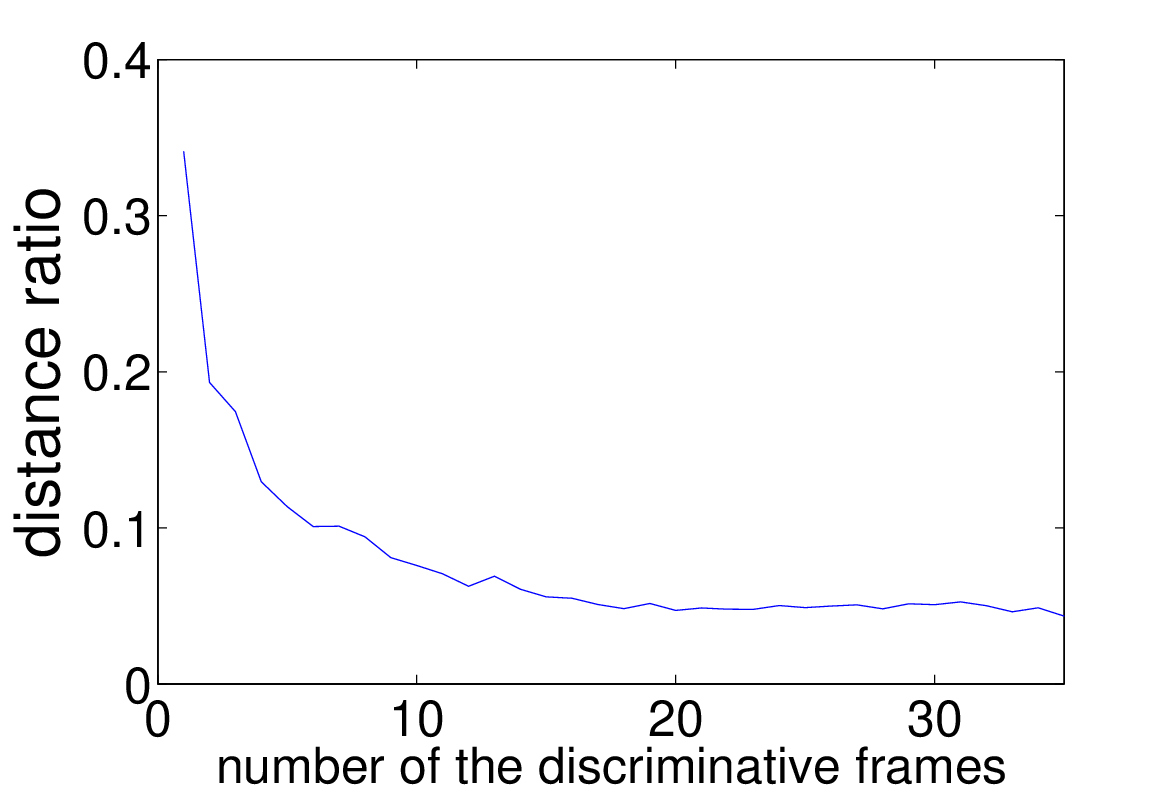}}
\subfigure[]{\includegraphics[width=0.33\textwidth]{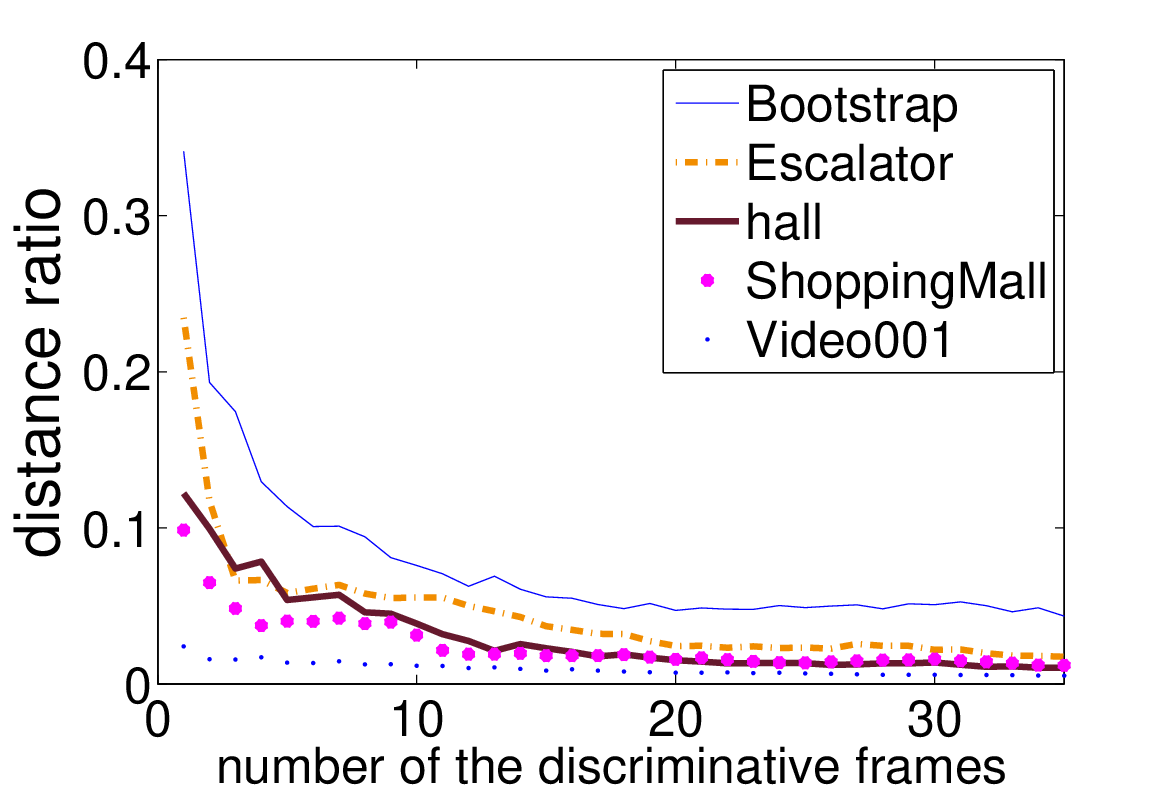}}
\caption{\label{Fig.5}The relationship between the number and the performance (1). The distance ratio is to divide the distance between the result and the standard by that between the standard and the original of coordinate.}
\end{center}
\end{figure}

First, we provide the details of our experiment on the "Bootstrap" sequence of the I2R dataset. A scene from this video takes place in front of a buffet. We use the first 300 frames in the sequence as our original frame set $\mathbf{D}$, and measure the performance of the SOIR algorithm when the number of frames of the selected discriminative frame set $\widetilde{\mathbf{D}}$ varies from 1 to 30. For comparison, we need a standard background, which we use all 300 frames to extract.

The results are shown in Fig. \ref{Fig.4}. In the figure, we can see that most of the extracted backgrounds are quite similar to the standard, even when the number of frames is small. However, it is a little disappointing   that the counter is not recovered exactly, even in our standard background. The two small fuzzy areas are the spaces just in front of the buffet, where people are continuously standing and taking the meal in nearly every frame. We measure the relationship between the rate of convergence and the number of discriminative frames, the results of which are shown in Fig. \ref{Fig.5}(a).

Next, we repeat the operations on some additional video clips, i.e., the "Escalator", "hall", and "ShoppingMall" sequences in the I2R dataset, and a real video sequence in the BMC dataset. The results of each video sequences (including "Bootstrap") are shown in Fig. \ref{Fig.5}(b).

We can see from Fig. \ref{Fig.5} that the performance is not very high when the number of frames is small. However, it improves as the number of frames  increases. When the number of frames is larger than 20, the ratio tends to become stable. The small fluctuation of each curve is caused by the non-background information of each new frame; however, the influence of this weakens after the frame is processed using our algorithm. In addition, we also find that the content of the video affects the results. In the "Video001" sequence, the foreground region is small. Thus, a small amount of frames already carry sufficient background information, and the curve is smooth. However, in the "Bootstrap" sequence, we use many more frames to deal with the changes in illumination, although the 30 results shown in Fig. \ref{Fig.4} all look the same.

Finally, we also explore the appropriate number of discriminative frames when the original frame number is larger, the results of which are shown in Fig. \ref{Fig.6}. We can see that the curve varies when the original number of frames increases because the discriminative frames are different. However, the distance ratio tends to become stable in all experiments when the number of discriminative frames increases to around 35. The ratio will improve if we use more frames, but the effect is unremarkable. This means that 35 frames or so already carry sufficient background information in most cases. If the content of the video sequence is pretty simple, fewer frames will be required, and oppositely, more frames will be needed if the content is complex.

\subsection{Experiments on the time consumption}
\begin{figure*}[!htb]
\centering
\includegraphics[width=0.9\textwidth]{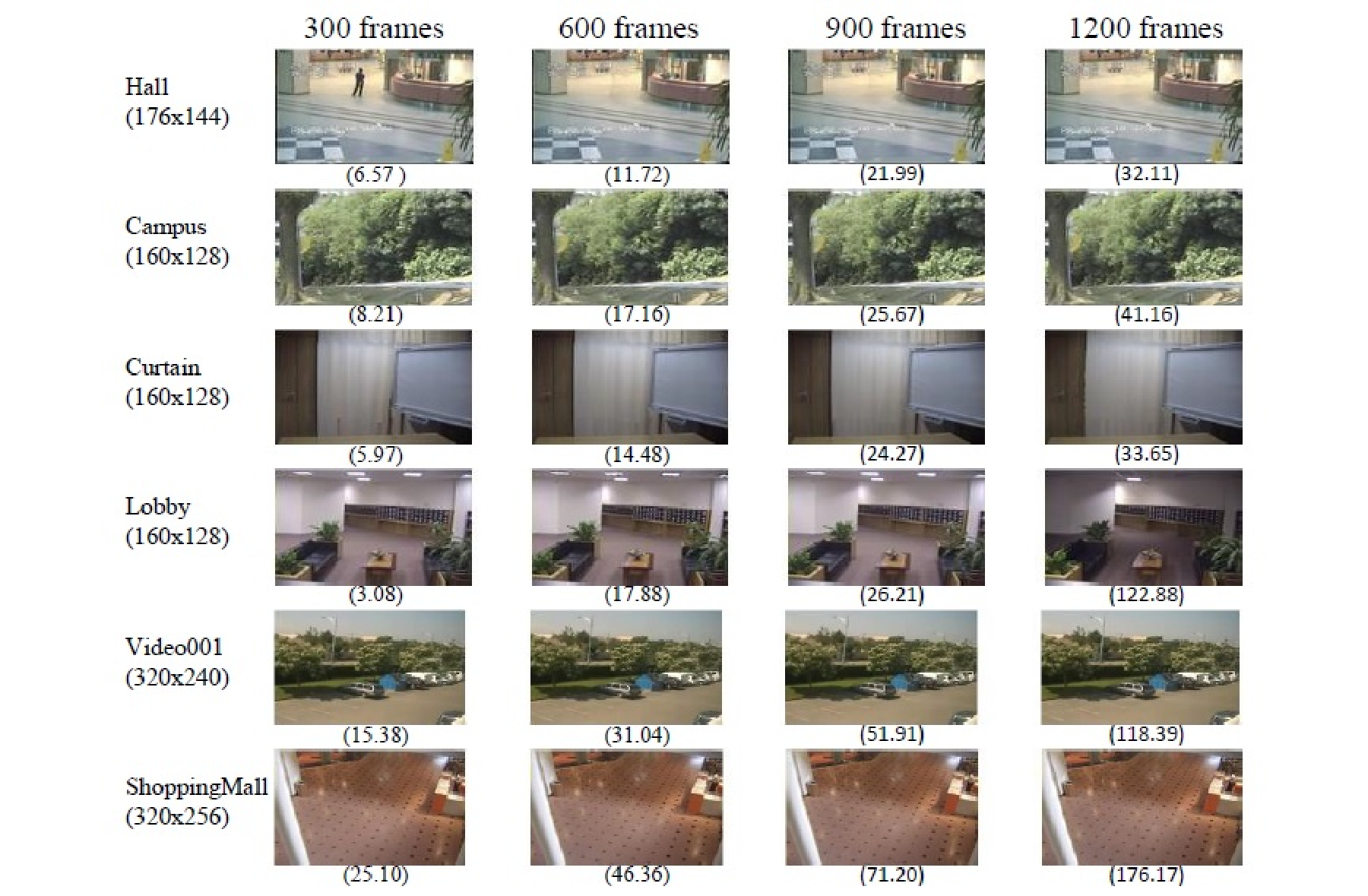}
\caption{\label{Fig.7}The experiments on different video sequences. The results are shown together with the time consumption(The unit of time is second). The sequences are from the I2R dataset and the BMC dataset.}
\end{figure*}

In this section, we discribe the time consumption of our model when solving tasks of different sizes. The majority of traditional methods are used for sequences with resolutions of around 150 $\times$ 150, and where the number of frames is usually around 50. When the scale of the data increases, these methods tend to become inefficient.

First, we focus on the number of frames of the sequence. We extract the background in four levels, i.e., from the first 300 frames, the first 600 frames, the first 900 frames and the first 1200 frames. The results are shown in Fig. \ref{Fig.7}.

\begin{figure*}[htb]
\centering
\includegraphics[width=0.8\textwidth]{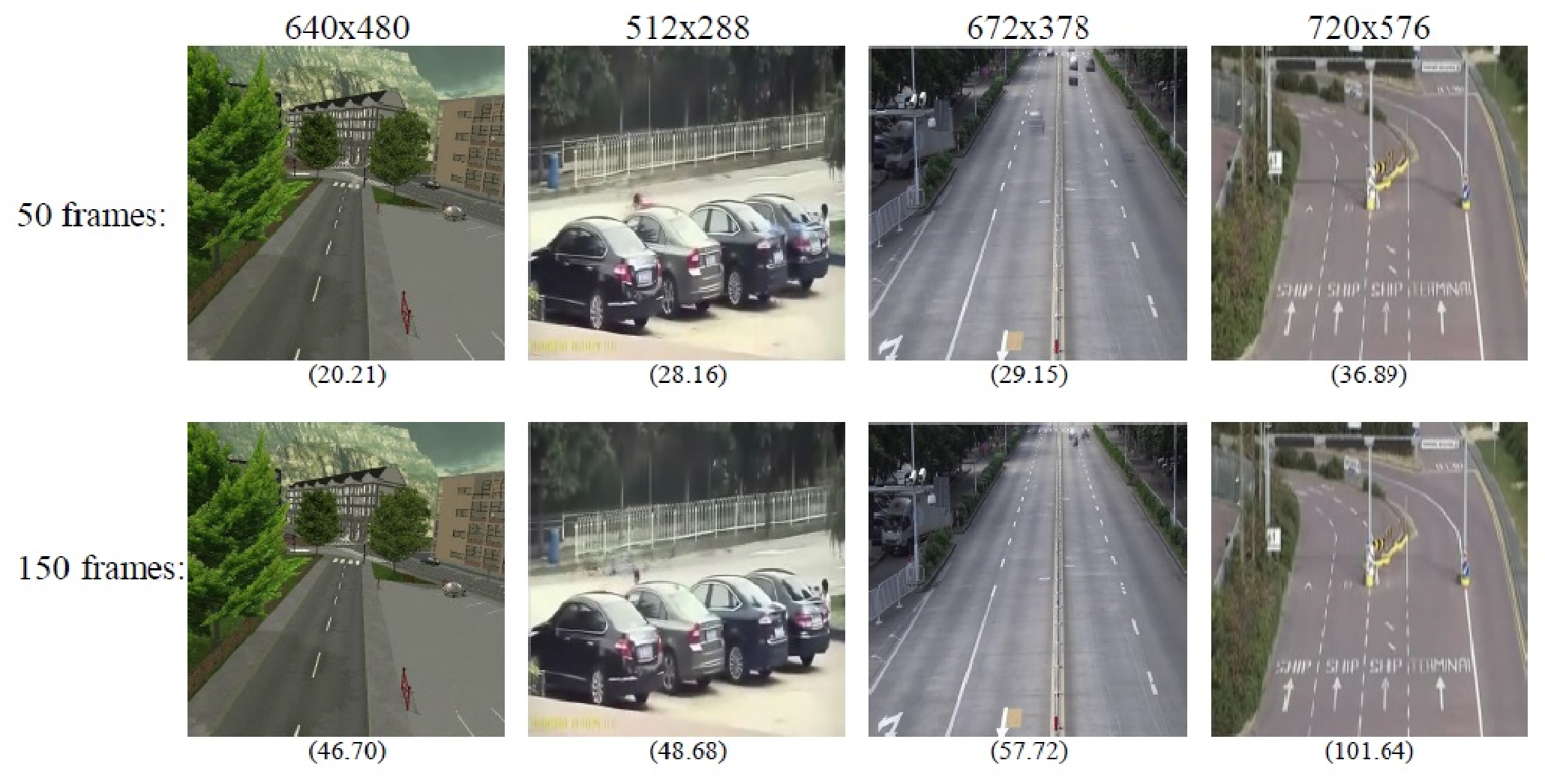}
\caption{\label{Fig.8}The experiments on the videos whose resolutions are much higher. The results are shown together with the time consumption (the unit of time is second). The resolutions are given out at the top of each column. The leftmost video is from BMC dataset while the others are from the Internet.}
\end{figure*}

When dealing with sequences with larger numbers of frames than usual, our model solves the background efficiently. Hundreds of frames only cost us dozens of seconds. As the number of frames increases, the precision of the extracted background is improved. Temporary static motion is a problem that exists in most traditional background modeling methods. Once a person  remains at a particular spot for a while, he may be considered a part of the background in a short video sequence. In our experiments, as the number of frames increases, the problem of temporary stay is perfectly solved, as illustrated in the results on the "Hall" sequence. We can also see that the time consumption is non-linear with the number of frames. On one hand, the uncertainty of a background created by the temporary stay may cost some additional time. On the other hand, the foreground content and noise also influence the time consumption.

Next, we test our model on video sequences of both low and high resolutions. We use four video sequences, the first of which is from the BMC dataset, and the other three are intersection monitoring video sequences from a public resource. We test our model on the first 50 frames and 150 frames of each sequence. The results are shown in Fig. \ref{Fig.8}.

The man-made video from the BMC dataset consumes the least amount of time. In the later three real-life videos, the time consumption increases as the resolution of each video sequence increases. Our model spends dozens of seconds solving the high-resolution video sequences. In addition, we can also conclude from Fig. \ref{Fig.8} that the content of the video sequence also influences the performance. In the third video, the distant cars move slowly in the fixed lens owing to the perspective. which is actually an approximation of temporary stay. We can see that 150 frames are still insufficient to resolve this phenomenon, and more frames are needed.

For comparison, we also examine the time consumption of some additional  methods. Because our model is a PCP model, two PCP-based methods are included, i.e., the Principal Component Pursuit (PCP) \cite{EXYJ2011} and the Detecting Contiguous Outliers in Low-Rank Representation (DECOLOR) \cite{XCW2013}, which perform well for small-scale problems. In addition, we also use the Single Gaussian (SG) model \cite{CATA1997}, which is currently the fastest method \cite{T2011}.

In the experiments, we use the "MovedObject" and "Bootstrap" sequences from the flowerwall dataset, and the "hall" and "Campus" sequences from the I2R dataset. For each video sequence, we use 150, 300, and 450 frames as the original frame set to extract the background. As shown in Table \ref{table1}, SG is the fastest, but its speed is maintained at the expense of accuracy, which is lower than that of most other popular methods \cite{T2011}.

\begin{table}[H]
\centering
\caption{\label{table1}The time consumption of PCP, DECOLOR, SG and SOIR.}
\begin{tabular}{ cccccc}
\hline
$\textrm{video}$  &$\textrm{number}$   &$\textrm{PCP}$  &$\textrm{DECOLOR}$  &$\textrm{SG}$ &$\textrm{SOIR}$\\
\hline
\hline
\multirow{3}{*}{$\textrm{MovedObject}$}  &$\textrm{150}$     &24.66      &43.76      &1.37 &2.65      \\
&$\textrm{300}$     &53.97      &74.66       &2.81 &6.19      \\
&$\textrm{450}$     &85.52      &124.12       &4.30 &11.90     \\
\hline
\multirow{3}{*}{$\textrm{Bootstrap}$}  &$\textrm{150}$     &61.27      &186.18       &1.58 &6.25      \\
&$\textrm{300}$     &124.61      &491.57       &3.43   &7.5      \\
&$\textrm{450}$     &207.56     &902.26       &5.43   &10.59      \\
\hline
\multirow{3}{*}{$\textrm{hall}$}  &$\textrm{150}$     &64.57      &144.28       &2.00 &3.75      \\
&$\textrm{300}$     &154.71      &303.98       &4.47 &5.94      \\
&$\textrm{450}$     &246.26      &599.91       &7.26  &9.61     \\
\hline
\multirow{3}{*}{$\textrm{Campus}$}  &$\textrm{150}$     &39.26      &27.18       &1.72 &5.66      \\
&$\textrm{300}$     &89.01      &47.13       &3.63 &6.94      \\
&$\textrm{450}$     &150.67      &80.14        &6.77 &12.09      \\
\hline
\end{tabular}
\end{table}

SOIR is on average ten times faster than other PCP-based models, and can almost reach the speed of SG. This level of speed is because the result of SOIR extracting the background from the discriminative frames, instead of the original frame set. When the scale of the data is large, the major time consumption of SOIR is in exploring these discriminative frames. Once they are obtained, however, we can model the background quickly and precisely. In the next section, we will illustrate how the accuracy of our method is high in dealing with real-life video sequences.

\subsection{Detecting the foreground}
\subsubsection{Evaluation on artificial dataset}

In this section, we provide the evaluation results obtained using the SABS dataset. Some approaches in the literatures \cite{ATAL2012,AHR2013} have been evaluated on the SABS dataset, and their recall-precision curves have been given. For a comparison of these curves, we also evaluate our algorithm on the SABS dataset, and give the corresponding recall-precision curves of our algorithm.

\begin{figure}[H]
\begin{center}
\includegraphics[width=0.4\textwidth]{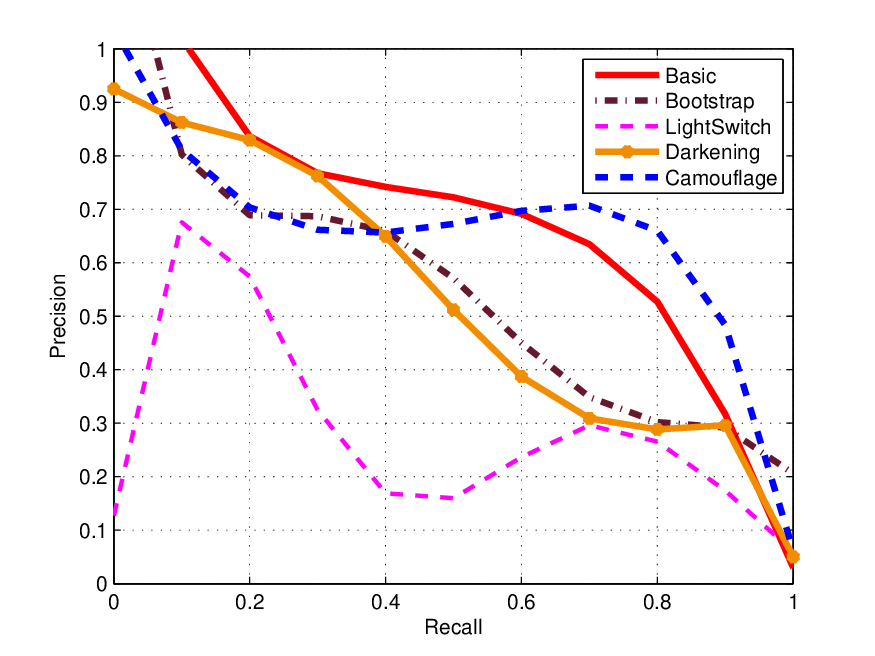}
\caption{\label{Fig.9}Precision and recall of our method for the videos in the SABS dataset.}
\end{center}
\end{figure}

In Fig. \ref{Fig.9}, the curves are evaluated on different scenes in the SABS dataset. Our method performs well for most scenes, especially in the "Basic" and "Camouflage" sequences. The performance for the "LightSwitch" sequence is not very good. As shown in the literature \cite{ATAL2012,AHR2013}, the light switch in the video is a huge challenge for most foreground detection methods.

\subsubsection{Evaluation on real scenes}
\begin{figure*}[htb]
\centering
\includegraphics[width=0.7\textwidth]{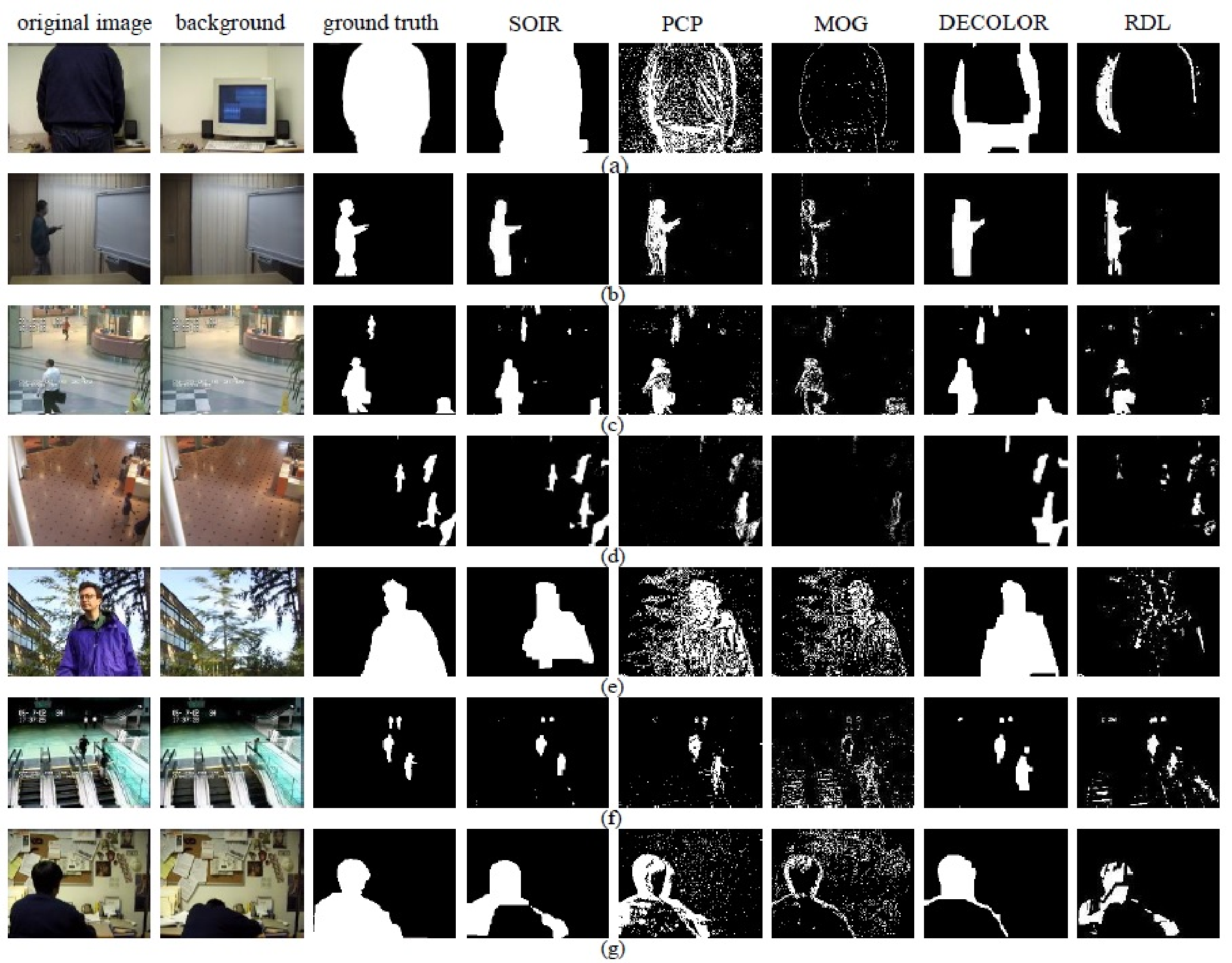}
\caption{\label{Fig.10}Results of detecting the foreground. From left to right: original image, exacted background by SOIR, ground truth, SOIR, PCP, MOG, DECOLOR,RDL. From top to bottom: Camouflage (b00251, flowerwall), Curtain (Curtain22772, I2R), hall (airport2180, I2R), ShoppingMall (ShoppingMall1980, I2R), WavingTrees (b00247, flowerwall), Escalator (airport4595, I2R), ForegroundAperture (b00489, flowerwall). }
\end{figure*}

We compare our performance with some other researches, i.e., Multiple of Gaussian (MOG) \cite{CW1999}, PCP \cite{EXYJ2011}, DECOLOR \cite{XCW2013}, and Robust Dictionary Learning (RDL) \cite{CXW2011}. Here, the fastest SG is not included, because its accuracy is lower than those of most popular methods \cite{T2011}. Thus, we use a more complex Gaussian model, i.e., MOG. We use the video sequences from the I2R and flowerwall datasets and compare the detected foreground region with the given hand-segmented foreground region. The test frame is chosen randomly from all hand-segmented frames. To avoid the influence of a temporary stay, we use 250 frames, the last of which is the test frame.

The sequences and results are shown in Fig. 10. In the experiments, SOIR can extract the background exactly for almost all of the sequences, and is robust to noise. In video sequence (g), the man remains at the same spot in all 250 frames.  We can see that our algorithm is robust to noise and performs well in foreground detection, which benefits from the accurate results of the background and the MRF model. DECOLOR also performs well because it also models the foreground using the MRF model. In most sequences, the results of SOIR are better than those of DECOLOR, because the extracted backgrounds of our algorithm are more exact.

To quantitatively evaluate the performance of the different algorithms, we compute the F-measure, which is derived from the precision and recall, and is computed as
\begin{equation}\label{eq35}
\mathop{\textrm{F-measure}}=\frac{2\times\mathop{\textrm{precision}}\times\mathop{\textrm{recall}}} {\mathop{\textrm{precision}}+\mathop{\textrm{recall}}}.
\end{equation}

\begin{table}[H]
\centering
\caption{\label{table2}F-Measures of the Sequences Shown in Fig.10 .}
\begin{tabular}{ ccccccc }
\hline
$\textrm{Sequence}$   &$\textrm{SOIR}$  &$\textrm{PCP}$  &$\textrm{MOG}$  &$\textrm{DECOLOR}$ &$\textrm{RDL}$\\
\hline
$\textrm{(a)}$     &$\textbf{0.9737}$      &0.6110       &0.2047       &0.5669     &0.1170   \\
$\textrm{(b)}$     &$\textbf{0.9020}$      &0.7129       &0.3841       &0.8244     &0.7326    \\
$\textrm{(c)}$     &$\textbf{0.8452}$      &0.6986       &0.5406       &0.7225     &0.6160    \\
$\textrm{(d)}$     &$\textbf{0.8314}$      &0.5248       &0.2498       &0.6439     &0.5367    \\
$\textrm{(e)}$     &0.8170      &0.6046       &0.4014       &$\textbf{0.8966}$     &0.3476    \\
$\textrm{(f)}$     &$\textbf{0.7972}$      &0.5902       &0.2455       &0.6487     &0.2399    \\
$\textrm{(g)}$     &$\textbf{0.6382}$      &0.5104       &0.1962       &0.3941     &0.5409    \\
\hline
\end{tabular}
\end{table}

Table \ref{table2} shows the F-measures of all detected foreground regions in Fig. 10. We can see that the results of SOIR are better than those of the other four methods for six sequences, i.e., (a),(b),(c),(d),(f), and (g). However, it is a little worse than DECOLOR in sequence (e). We can also see that the performance of SOIR varies among the different video sequences. On one hand, this is due to the result of the background extraction, which is the case for sequence (g). On the other hand, the instability of the background also affects the performance of SOIR.

\section{Conclusion and future work}
\label{sec6}

In this paper, we propose a Sparse Outlier Iterative Removal (SOIR) algorithm to model the background of a video sequence. We find that a few discriminative frames are already sufficient to model the background. We use the sparse representation process to reduce the size of the video. Although it takes our algorithm some time to explore the 'discriminative' frames, it saves much more time in modeling the background. A cyclic iteration process is proposed for background extraction. SOIR achieves both high accuracy and high speed simultaneously when dealing with real-life video sequences. In particular, SOIR has an advantage in solving large-scale tasks.

As future work, we will deal with some additional complex problems in which the background is no longer stable among different frames.


%



\ifCLASSOPTIONcaptionsoff
  \newpage
\fi




\begin{thebibliography}{40}


\bibitem{HCQ2011} H. Li, C. Shen, Q. Shi, Real-time visual tracking using compressive sensing, Proc. IEEE Int. Conf. Comput. Vision Pattern Recognit., Providence, RI, USA, Jun. 2011, pp. 1305-1312.

\bibitem{KLM2012} K. Zhang, L. Zhang, M. Yang, Real-time compressive tracking, European Conf. on Comput. Vision,  Oct. 2012, pp. 864-877.


\bibitem{T2011} T. Bouwmans, Recent advanced statical background modeling for foreground detection-A systematic survey, Recent Patents on Comput. Sci., vol. 6, no. 3, pp. 147-176, Nov. 2011.

\bibitem{AOM2006} C. Kim, J. Hwang, Object-based video abstraction for video surveillance systems, IEEE Trans. Circuits Syst. Video Technol., vol. 12, no. 12, pp. 1128-1138, Apr. 2010.

\bibitem{SHL2012} S. Li, H. Lu, L. Zhang, Arbitrary body segmentation in static images, Pattern Recognit., vol. 45, no. 9, pp. 3402-3413, Sep. 2012.

\bibitem{DH2013} D. Park, H. Byun, A unified approach to background adaptation and initialization in public scenes,  Pattern Recognit., vol. 46, no. 7, pp. 1985-1997, Jul. 2013.

\bibitem{WJW2008}  W. Wang, J. Yang, W. Gao, Modeling background and segmenting moving objects from compressed video, IEEE Trans. Circuits Syst. Video Technol., vol. 18, no. 5, pp. 670-681, May. 2008.

\bibitem{MO2012} M. Droogenbroeck, O. Paquot, Background subtraction: Experiments and improvements for vibe, IEEE Int. Conf. Comput. Vision Pattern Recognit. Workshop, Providence, RI, USA, Jun. 2012, pp. 32-37.

\bibitem{VN2008} V. Mahadevan, N. Vasconcelos, Background subtraction in highly dynamic scenes, Proc. IEEE Int. Conf. Comput. Vision Pattern Recognit., horage, AK, USA, Jun. 2008, pp. 1-6.

\bibitem{YCCY2007} Y. Chen, C. Chen, C. Huang, Y. Hung, Efficient hierarchical method for background subtraction, Pattern Recognit., vol. 40, no. 10, pp. 2706-2715, Oct. 2007.


\bibitem{STS2011}  S. Wang, T. Su, S. Lai, Detecting moving objects from dynamic background with shadow removal, IEEE Int. Conf. Acoust. Speech and Signal Process., Prague, CZ, May. 2011, pp. 925-928.

\bibitem{AT2008} A. Ulges, T. Breuel, A local discriminative model for background subtraction, Pattern Recognit., Springer Berlin Heidelberg, Jun. 2008, pp. 507-516.

\bibitem{CATA1997} C. Wren, A. Azarbayejani, T. Darrell, A. Pentland, Pfinder:Real-time tracking of human body, IEEE Trans. Pattern Anal. Mach. Intell., vol. 19, no. 7, pp. 780-785, Jul. 1997.

\bibitem{CW1999} C. Stauffer, W. Grimson, Adaptive background mixture models for real-time tracking, Proc. IEEE Int. Conf. Comput. Vision Pattern Recognit., Fort Collins, CO, USA, 1999.


\bibitem{D2005} D. Lee, Effective Gaussian mixture learning for video background subtraction, IEEE Trans. Pattern Anal. and Mach. Intell., vol. 27, no. 5, pp. 827-832, May. 2005.

\bibitem{JHGJ2011} J. K. Suhr, H. G. Jung, G. Li, J. Kim, Mixture of Gaussians-Based Background Subtraction for Bayer-Pattern Image Sequences, IEEE Trans. Circuits Syst. Video Technol., vol. 21, no. 3, pp. 365-370, Mar. 2011.

\bibitem{HJT2011} H. Lin, J. Chuang, T.Liu, Regularized background adaptation: a novel learning rate control scheme for Gaussian mixture modeling, IEEE Trans. Image Process., vol. 20, no. 3, pp. 822-836, 2011.


\bibitem{JYCC2011} J. Guo, Y. Liu, C. Hsia, C. Hsu, Hierarchical method for foreground detection using codebook model, IEEE Trans. Circuits Syst. Video Technol., vol. 21, no. 6, pp. 804-815, Jun. 2011.

\bibitem{AM2011}A. Zaharescu, M. Jamieson, Multi-scale multi-feature codebook-based background subtraction, IEEE Int. Conf. Comput. Vision Workshop, Barcelona, Spain, Nov. 2011, pp. 1753-1760.

\bibitem{AN2012} A. Hamad, N. Tsumura, Background subtraction based on time-series clustering and statistical modeling, Optical Review, vol. 19, no. 2, pp. 110-120, Mar. 2012.

\bibitem{ADL2000} A. Elgammal, D. Harwood, L. Davis. Non-parametric model for background subtraction, European Conf. on Comput. Vision, Springer Berlin Heidelberg, Jun. 2000, pp. 751-767.

\bibitem{EMA2012} E. Learned-Miller, M. Narayana, A. Hanson, Background modeling using adaptive pixelwise kernel variances in a hybrid feature space, IEEE Int. Conf. Comput. Vision Pattern Recognit., Providence, RI, USA, Jun. 2012, pp. 2104-2111.


\bibitem{OM2011} O. Barnich, M. Van Droogenbroeck, ViBe: A universal background subtraction algorithm for video sequences, IEEE Trans. Image Process., vol. 20, no. 6, pp. 1709-1724, Jun. 2011.


\bibitem{LWIQ2004}  L. Li, W. Huang, I. Gu , Q. Tians Statistical modeling of complex backgrounds for foreground object detections, IEEE Trans Image Process., vol. 13, no. 11, pp. 1459-1472, 2004.

\bibitem{LWXG2010}  L. Zhang, W. Dong, X. Wu, G. Shi, Spatial-Temporal color video reconstruction from noisy CFA sequence, IEEE Trans. Circuits Sys. Video Technol., vol. 20, no. 6, pp. 838-847, Jun. 2010.

\bibitem{SGV2010} S. Liao, G. Zhao, V. Kellokumpu, M. Pietikainen, S. Li, Modeling pixel process with scale invariant local patterns for background subtraction in complex scenes, IEEE Int. Conf. Comput. Vision Pattern Recognit., San Francisco, CA, USA, Jun. 2010, pp. 1301-1306.

\bibitem{NBA2000} N. Oliver, B. Rosario, A. Pentland, A bayesian computer vision system for modeling human interactions, IEEE Trans. Pattern Anal. Mach. Intell., vol. 22, no. 8, pp. 831-843, Aug. 2000.

\bibitem{ZXJEY2010}  Z. Zhou, X. Li, J. Wright, E. Candes, Y. Ma, Stable principal component pursuit, IEEE Int. Symp. Inf. Theory Pro., Austin, TX, USA, Jun. 2010, pp. 1518-1522.

\bibitem{EXYJ2011} E. Candes, X. Li, Y. Ma, and J. Wright, Robust principal component analysis? J. ACM, vol. 58, no. 3, May. 2011.

\bibitem{GA2011}  G. Tang, A. Nehorai, Robust principal component analysis based on low-rank and block-sparse matrix decomposition, IEEE Annu. Conf. Inf. Sci. Syst., Baltimore, MD, USA, Mar. 2011, pp. 1-5.

\bibitem{CTE2012} C. Guyon, T. Bouwmans, E. Zahzah, Foreground detection based on low-rank and block-sparse matrix decomposition, IEEE Int. Conf. Image Process., Orlando, FL, USA, Oct. 2012, pp. 1225-1228.

\bibitem{YDJ2011} Y. Xu, D. Zhang, J. Yang. A two-phase test sample sparse representation method for use with face recognition. IEEE Trans. Circuits Syst. Video Technol., vol. 21, no.9, pp. 1255-1262, Apr. 2011.

\bibitem{XCW2013} X. Zhou, C. Yang, W. Yu, Moving object detection by detecting contiguous outliers in the low-Rank representation, IEEE Trans. Pattern Anal. Mach. Intell., vol. 35, no. 3, pp. 597-610, Mar. 2013.

\bibitem{JAASY2009} J. Wright, A. Yang,  A. Ganesh, S. Sastry, Y. Ma, Robust face recognition via sparse representation, IEEE Trans. Pattern Anal. Mach. Intell., vol. 31, no. 2, pp. 210-227, Feb. 2009.

\bibitem{EGR2012} E. Elhamifar, G. Sapiro, R. Vidal, See all by looking at a few: Sparse modeling for finding representative objects, IEEE Conf. Comput. Vision Pattern Recognit., Providence, RI, USA, Jun. 2012, pp. 1600-1607.

\bibitem{WDLXD2013} W. Zuo, D. Meng, L. Zhang, X. Feng, D. Zhang, A generalized iterated shrinkage algorithm for non-convex sparse coding, IEEE Int. Conf. Comput. Vision, Sydney, Australia, 2013.



\bibitem{JYW2009}  J.Yang, Y.Wang, W.Xu. Image and video denoising using adaptive dual-tree discrete wavelet packets. IEEE Trans. Circuits Syst. Video
Technol., vol. 19, no. 5, pp. 642-655, Mar. 2009.

\bibitem{RAM2011} R.Sivalingam, A.D'Souza, M.Bazakos. Dictionary learning for robust background modeling, IEEE Int. Conf. Robotics and automation, shanghai, China, May. 2011, pp. 4234-4239.

\bibitem{CXW2011} C. Zhao, X. Wang , W. Cham, Background subtraction via robust dictionary learning, J. Image Video Process., Feb. 2011.


\bibitem{BT2006}  B. Bader, T. Kolda, Algorithm 862: MATLAB tensor classes for fast algorithm prototyping, ACM Trans. Math. Software, vol. 32, no. 4, pp. 635-653, Dec. 2006.

\bibitem{TB2009} T. Kolda, B. Bader, Tensor decompositions and applications, Soc. Ind. Appl. Math., vol. 51, no. 3, pp. 455-500, 2009.

\bibitem{J2006} J. Tropp, Algorithms for simultaneous sparse approximation. Part II: Convex relaxation, IEEE Trans. Signal Process., vol. 86, no. 4, pp. 589-602, Mar. 2006.

\bibitem{XJ2013}  X. Yuan and J. Yang, Sparse and low-rank matrix decomposition via alternating direction methods, Pacific J. Optimization, vol. 9, no. 1, pp. 167-180, 2013.

\bibitem{ZRZ2011}   Z. Lin, R. Liu, and Z. Su, Linearized alternating direction method with adaptive penalty for low rank representation, Neural Inf. Process. Syst., Dec. 2011.

\bibitem{JFJG2009} J. Mairal, F. Bach , J. Ponce , G. Sapiro, Online dictionary learning for sparse coding, ACM Annu. Int. Conf. Mach. Learning, NY, USA, 2009, pp. 689-696.

\bibitem{RW2012} R.Johnsonbaugh, W.Pfaffenberger, Foundations of mathematical analysis, Courier Dover Publications, 2012.


\bibitem{S2009} S. Li, Markov random field modeling in image analysis, Springer Publishing Company, 2009.

\bibitem{CH2010} C. Chung, H. Chen, Video object extraction via MRF-based contour tracking, IEEE Trans. Circuits Syst. Video Tech., vol. 20, no. 1, pp. 149-155, Jan. 2010.

\bibitem{YOR2001} Y. Boykov, O. Veksler , R. Zabih, Fast approximate energy minimization via graph cuts, IEEE Tran. Pattern Anal. Mach. Intell., vol. 23, no. 11, pp. 1222-1239, Nov. 2001.

\bibitem{KJBB1999} K. Toyama, J. Krumm, B. Brumitt, B. Meyers, Wallflower: Principles and practice of background maintenance, IEEE Int. Conf. Comput. Vision, Kerkyra, Greece, Sep. 1999, pp. 255-261.

\bibitem{SBG2013} S. Brutzer, B. Hoferlin, G. Heidemann, Evaluation of background subtraction techniques for video surveillance, IEEE Int. Conf. Comput. Vision Pattern Recognit., Providence, RI, USA, Jun. 2013, pp. 1937-1944.


\bibitem{ATAL2012} A. Vacavant, T. Chateau, A. Wilhelm, L. Lequievre, A benchmark dataset for outdoor foreground/background extraction,  Asian Conf. Comput. Vision Workshop, Daejeon, Korea, Nov. 2012, pp. 291-300.


\bibitem{AHR2013} A. Shimada, H. Nagahara, R. Taniguchi, Background modeling based on bidirectional analysis, IEEE Int. Conf. Comput. Vision Pattern Recognit., Providence, RI, USA, Jun. 2013, pp. 1979-1986.



\end{thebibliography}
%

%

\begin{IEEEbiography}[{\includegraphics[width=1in,height=1.25in,clip,keepaspectratio]{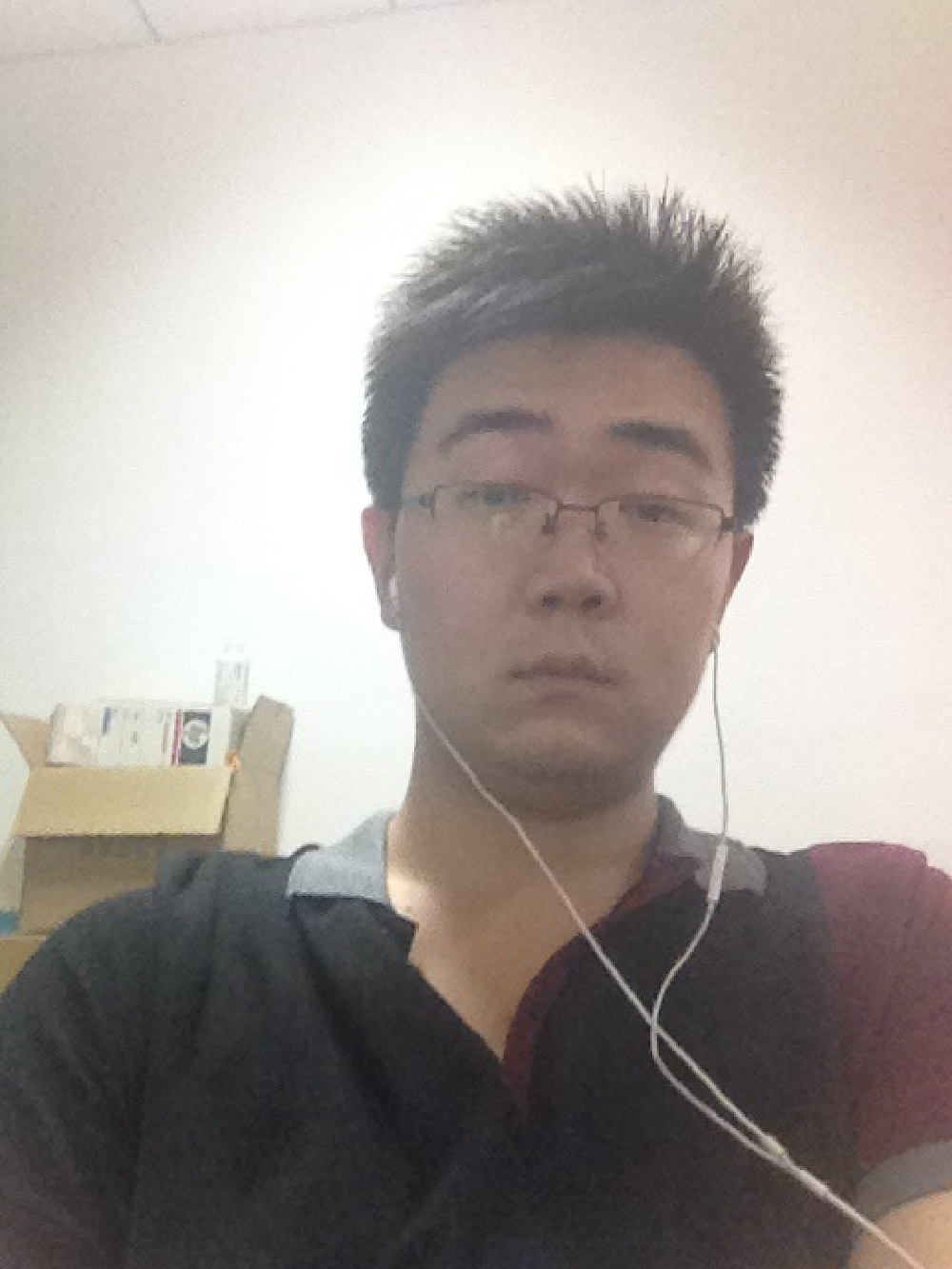}}]{Lihao Li}
was born in 1989. He received his B.S. degree in Applied Mathematics from Tianjin University in 2012 and his M.S. degree in Computational Mathematics from Tianjin University in 2014. Now he is a Ph.D. student in school of Computer Science and Technology, Tianjin University. His research interests focus on low-rank matrix recovery, machine learning, background modeling and sparse coding.
\end{IEEEbiography}

\begin{IEEEbiography}[{\includegraphics[width=1in,height=1.25in,clip,keepaspectratio]{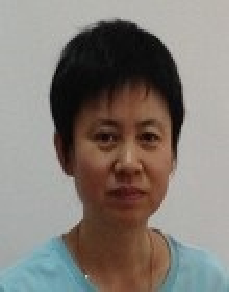}}]{Ping Wang}
was born in 1967. She received her B.S., M.S. and Ph.D. degree from Tianjin University, in 1988, 1991, and 1998, respectively. Now she is a professor, master¡¯s supervisor and Ph.D. supervisor in the Mathematics Department, Tianjin University. She is also a member of CCF (E200011015M) and ACM (4157113). Her research interests include signal and information processing, pattern recognition and image processing.
\end{IEEEbiography}
\vfill

\begin{IEEEbiography}[{\includegraphics[width=1in,height=1.25in,clip,keepaspectratio]{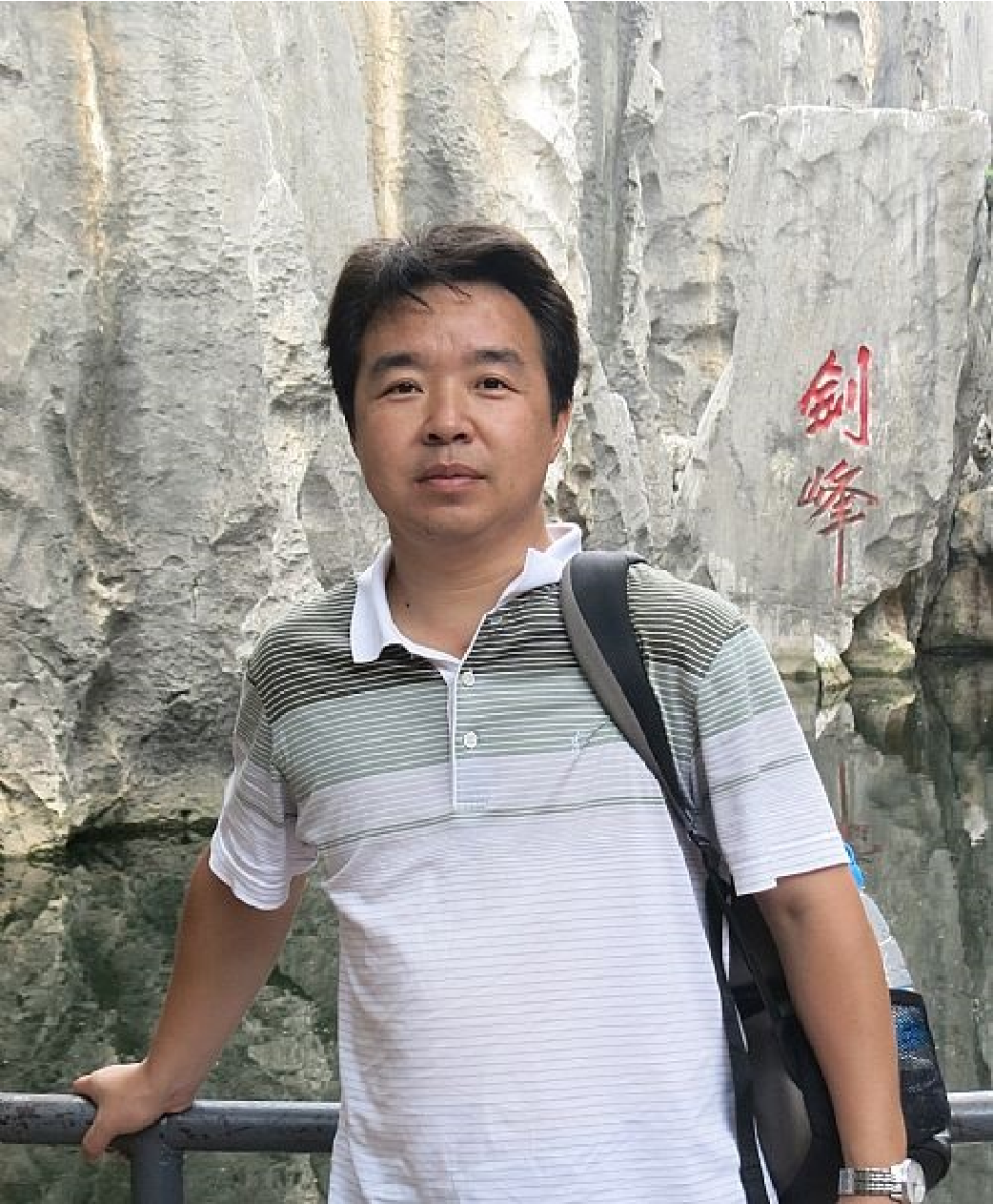}}]{Qinghua Hu}
received his B. S., M. S. and Ph.D. degrees from Harbin Institute of Technology, Harbin, China in 1999, 2002 and 2008, respectively. He worked as a Postdoctoral Fellow with Department of Computing, Hong Kong Polytechnic University from 2009 to 2011. Now he is a full professor and vice dean of the school of Computer Science and Technology, Tianjin University. His research interests are focused on rough sets, granular computing, data mining for classification and regression. He once acted as PC Co-Chairs of RSCTC 2010, CRSSC 2012, 2014, RSKT 2014 and ICMLC 2014, and severs as referee for a great number of journals and conferences. He has published more than 100 journal and conference papers in the areas of granular computing based machine learning, reasoning with uncertainty, pattern recognition and fault diagnosis.
\end{IEEEbiography}

\begin{IEEEbiography}[{\includegraphics[width=1in,height=1.25in,clip,keepaspectratio]{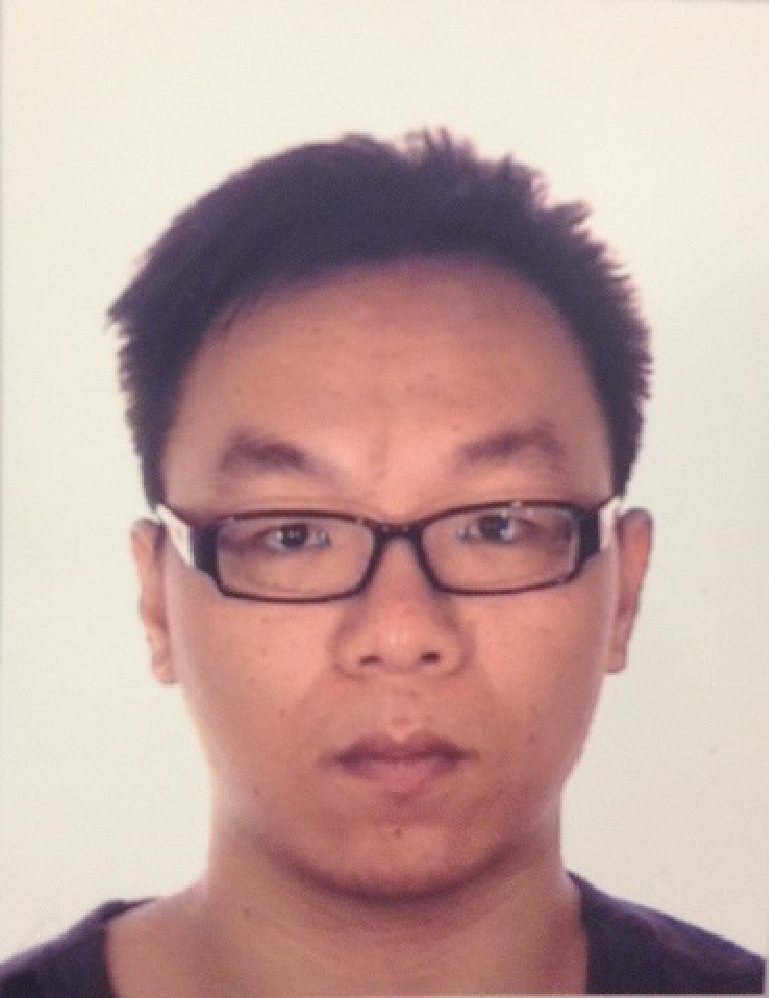}}]{Sijia Cai}
was born in 1988. He received his B.S. degree in Applied Mathematics from Tianjin University in 2011 and his M.S. degree in Computational Mathematics from Tianjin University in 2014. Now, He is an exchange student in Hong Kong Polytechnic University. His main research interests include sparse optimization, multi-linear analysis, image processing and machine learning.
\end{IEEEbiography}
\vfill





\end{document}